\documentclass[sigconf]{acmart}
\usepackage{colortbl}
\usepackage{xcolor}
\usepackage{arydshln}
\usepackage{multirow}
\usepackage{subcaption}
\definecolor{Gray}{gray}{0.85}
\definecolor{red}{cmyk}{0,1,1,.098}
\definecolor{blue}{cmyk}{1,1,0,.1}

\AtBeginDocument{%
  \providecommand\BibTeX{{%
    \normalfont B\kern-0.5em{\scshape i\kern-0.25em b}\kern-0.8em\TeX}}}

\copyrightyear{2020}
\acmYear{2020}
\setcopyright{acmcopyright}
\acmConference[HRI '20]{Proceedings of the 2020 ACM/IEEE International Conference on Human-Robot Interaction}{March 23--26, 2020}{Cambridge, United Kingdom}
\acmBooktitle{Proceedings of the 2020 ACM/IEEE International Conference on Human-Robot Interaction (HRI '20), March 23--26, 2020, Cambridge, United Kingdom}
\acmPrice{15.00}
\acmDOI{10.1145/3319502.3374800}
\acmISBN{978-1-4503-6746-2/20/03}

\begin{document}

\title[A Robot by Any Other Frame]{A Robot by Any Other Frame: Framing and Behaviour Influence Mind Perception in Virtual but not Real-World Environments}

\author{Sebastian Wallk\"otter}
\email{sebastian.wallkotter@it.uu.se}
\affiliation{%
  \institution{Uppsala University}
  \streetaddress{Box 337}
  \city{Uppsala}
  \state{Sweden}
  \postcode{751 05}
}

\author{Rebecca Stower}
\affiliation{%
  \institution{Jacobs University}
  \streetaddress{Campus Ring 1, Research IV}
  \city{Bremen}
  \state{Germany}
  \postcode{28759}
}

\author{Arvid Kappas}
\affiliation{%
  \institution{Jacobs University}
  \streetaddress{Campus Ring 1, Research IV}
  \city{Bremen}
  \state{Germany}
  \postcode{28759}
}

\author{Ginevra Castellano}
\affiliation{%
  \institution{Uppsala University}
  \streetaddress{Box 337}
  \city{Uppsala}
  \state{Sweden}
  \postcode{751 05}
}

\renewcommand{\shortauthors}{Wallk\"otter, et al.}

\begin{abstract}
Mind perception in robots has been an understudied construct in human-robot interaction (HRI) compared to similar concepts such as anthropomorphism and the intentional stance. In a series of three experiments, we identify two factors that could potentially influence mind perception and moral concern in robots: how the robot is introduced (framing), and how the robot acts (social behaviour). In the first two \textbf{online} experiments, we show that both framing and behaviour independently influence participants' mind perception. However, when we combined both variables in the following \textbf{real-world} experiment, these effects failed to replicate. We hence identify a third factor post-hoc: the online versus real-world nature of the interactions. After analysing potential confounds, we tentatively suggest that mind perception is harder to influence in real-world experiments, as manipulations are harder to isolate compared to virtual experiments, which only provide a slice of the interaction.
\end{abstract}



\maketitle

\section{Introduction}\label{sec:intro}

\begin{figure}[t]
  \includegraphics[width=\linewidth]{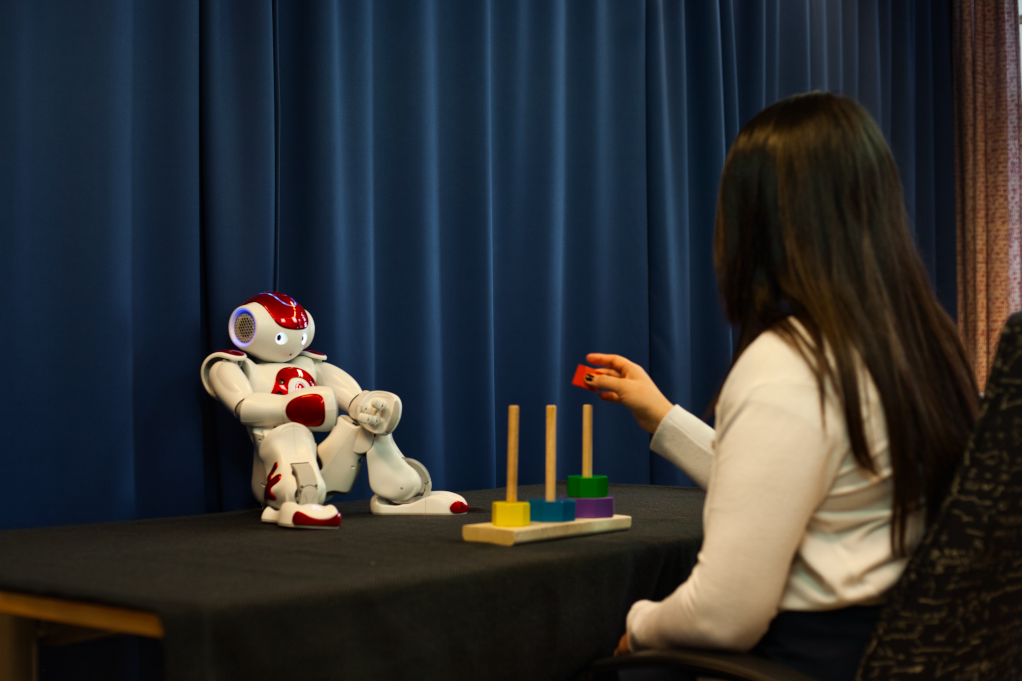}
  \caption{A NAO robot collaborating with a user to solve the Tower of Hanoi.}
  \label{fig:teaser}
\end{figure}

Within the human-robot-interaction (HRI) community there is growing acceptance of the idea that humans treat robots as social agents \cite{Graaf2019}, apply social norms to robots \cite{Vanman2019}, and, in some circumstances, treat robots as moral agents \cite{Malle2016}. All of these are related to the concept of mind perception, or how much agency a robot is seen to have \cite{Gray2007, Abubshait2017}. However, whilst the attribution of a mind to a robot may at times be desirable \cite{Wiese2017}, a mismatch between a robot's perceived mind and actual mind, i.e., its true capabilities, could prove detrimental for a successful interaction \cite{Wiese2017, Koda2016, Fink2012}. 

Investigating which factors actually influence mind perception in robots is, therefore, at the core of our research. We study two possible contributing factors: \textit{framing} (how the robot is introduced), and \textit{social behaviour} (e.g., speech and non-verbal cues provided by the robot). Work on anthropomorphism and mind perception in robots has largely focused on embodiment and appearance, with less emphasis on social behaviours \cite{Ghazali2019a, Bartneck2009, Fink2012}. Similarly, although the effect of framing on mind perception has been studied with virtual agents \cite{Caruana2017, Wiese2012, Wykowska2014}, this has yet to be replicated with embodied robots. As such, \textbf{the primary goal of this research is to investigate the interaction between framing and social behaviours on mind perception of a social robot}.

Further, given the recent surge of interest in ethical robotics\footnote{See for example: https://ec.europa.eu/digital-single-market/en/news/ethics-guidelines-trustworthy-ai} and explainable AI \cite{anjomshoae2019explainable}, it is also timely to ask to what extent mind perception contributes to this debate. Considering that mind perception has been linked to moral concern in robots \cite{Nomura2019a}, we go beyond just looking at mind perception, and investigate if the degree of moral concern attributed to robots \cite{Nomura2019} changes according to framing and behaviour.

We first conducted two experiments on Amazon Mechanical Turk (AMT) validating a framing manipulation (experiment 1), using an image of the robot and descriptive text, and a social behaviour manipulation (experiment 2), using a video of the robot's behaviour. Then, we conducted an experimental study in the real world to investigate the interaction between these factors. We hypothesise that robots which are framed with higher mind perception and present social behaviours will cause higher mind perception and be afforded more moral standing than low-mind frame, non-social robots. We also hypothesise that when the frame and behaviour are in conflict (e.g., high-mind frame, non-social behaviour) the behaviour of the robot will be more influential in determining mind perception and moral standing. Additionally, we predict a significant correlation between mind perception, anthropomorphism, and moral concern.

After looking at the results, we found a surprising lack of replication between the first two experiments, and the third. We, hence, discuss potential confounds, and, \textbf{as our second contribution, identify the nature of the interaction (real-world versus virtual) as the primary source for the lack of replication}.

\section{Related Work}\label{sec:relatedwork}
The HRI community has used various terminologies and constructs to refer to the attribution of human-like properties to robotic agents. Anthropomorphism is one of the most widely used terms, and refers to the general psychological phenomenon of attributing perceived human-like properties to non-human agents \cite{Epley2007, Zlotowski2015, Fink2012}. A related concept is the intentional stance, which focuses on explaining others' behaviour with reference to their mental states \cite{Schellen2019, Marchesi2019}. Finally, mind perception, as the focus of this paper, is related to the intentional stance \cite{Wiese2017, Schellen2019} and refers to the extent to which agents are seen as possessing a mind and agency \cite{Gray2007}. 

There is a general consensus that the more human-like the robot appears to be, the greater the degree of anthropomorphism \cite{Eyssel2012, Bartneck2009, Broadbent2013}. Several measures have been developed which target anthropomorphism, with one of the most popular being the Godspeed questionnaire \cite{Bartneck2009a}. As such, we find it also worthwhile to investigate how anthropomorphism relates to mind perception. 

The idea of mind perception was first introduced by \citet{Gray2007}. In their seminal study, they compared $11$ different agents (including a robot) on a variety of capabilities, such as pain, desire, or the ability to have goals. They found that mind perception could be divided into two dimensions: agency (the perceived ability to do things), and experience (the perceived ability to feel things). However, whilst a robot (Kismet \footnote{http://www.ai.mit.edu/projects/sociable/baby-bits.html}) was one of the agents used in their initial study, many different kinds of robots exist \cite{Phillips2018}, and there are still open questions as to how different properties of these robots affect ratings of mind perception.

Social behaviour is one such factor that could potentially influence mind perception. In the context of HRI, robots may be equipped to display a number of human-like behaviours, such as gaze following, joint attention, back-channelling, personalisation, feedback, and verbal and non-verbal cues \cite{Mutlu2006, Jung2013, Wills2016, Corrigan2013, Ahmad2019, Breazeal2005, Wigdor2016}. Robot behaviour can influence perceptions of machine or human-likeness \cite{Park2011, Fink2012}. \citet{Abubshait2017} also investigate the effect of behaviour on mind perception in virtual agents; however, focus on the reliability of the robot as a social behaviour. More complex social behaviours such as cheating \cite{Short2010} or making mistakes \cite{Salem2013, Mirnig2017} have otherwise been studied in the context of perceived human-likeness, with mixed findings. As such, given the complex interplay between robot errors, human performance, and mind perception, a baseline understanding of how social behaviours in robots influences mind perception is still needed. As there can be more variance in the behaviour of a given autonomous social robot than in its appearance, behaviour may be of equal or greater importance \cite{Wiese2017}. Therefore, we aim to address this gap, and investigate if and how behaviour can affect the perception of a mind in a social robot.

A second factor which could influence mind perception in social robots is how the robot is framed. Framing refers to the prior information a person has about the robot, such as prior expectations, beliefs, or knowledge \cite{Kwon2016}.  Within HRI, there is already some research which investigates how framing a robot prior to an interaction influences participants' subsequent judgements and behaviour \cite{Westlund2016, Stenzel2012, Groom2011, Rea2018, Rea2019, Thellman2017}. For virtual agents, several studies have investigated how framing an agent as being controlled by a human rather than a computer leads to greater attributions of mind \cite{Caruana2017, Wiese2012, Wykowska2014}. However, in addition to most of this research being conducted with virtual agents, rather than robots, these studies focus on primarily neuropsychological measures of mind perception, without validation of participants subjective experiences (but see \citet{Caruana2017} for an exception). As such, \textit{no research has yet directly investigated the effect of framing on mind perception in robots}. Hence, we investigate how two different frames, high and low mind, affect participant's mind perception in HRI.

A secondary element of the \citet{Gray2007} study involves the relation of mind perception to moral concern. They showed a link between the sub-dimensions of mind perception and perceiving an entity as moral agent, or moral patient respectively \cite{Gray2007}. Additionally, robots with differing embodiment and behaviour may be afforded different levels of moral concern \cite{Nomura2019a, Malle2016}. Further, framing can influence the expansion or reduction of people's moral inclusiveness \cite{Laham2009}; however, this link has yet to be explored in HRI. In light of this discussion, we chose to also include the recently proposed moral concern for robots scale \cite{Nomura2019} to investigate the relationship between mind perception and moral concern for robots.

\section{Technical Setup and Scenario}\label{sec:technical_spec}

\begin{figure}[t]
    \centering
    \includegraphics[width=\linewidth]{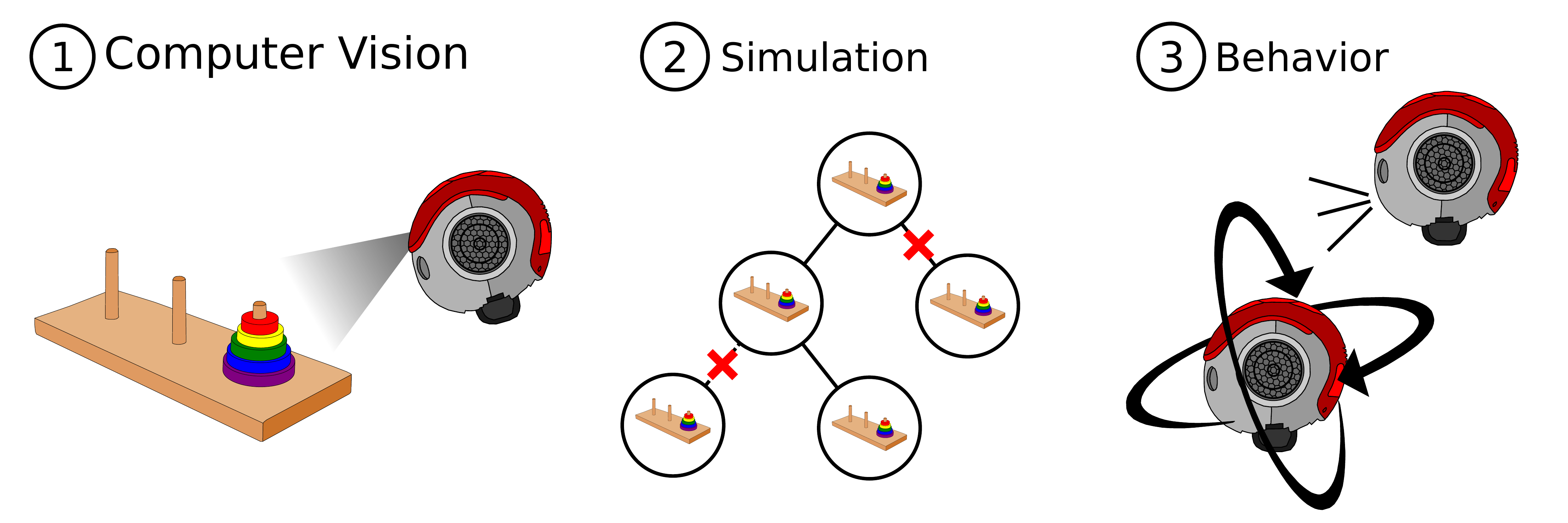}
    \caption{A graphic visualisation of the pipeline NAO uses.}
    \label{fig:technical_pipeline}
\end{figure}

For this research we programmed a Softbank Robotics NAOv5 robot (naoqi v2.1.4.13) to autonomously play the Tower of Hanoi with users. We modified the original Tower of Hanoi puzzle into a two player game by having the human and robot take turns in making a move. We chose to construct the interaction around the Tower of Hanoi, as it has been used previously to study aspects of HRI such as embodiment or (gaze) behaviour \cite{Tsiakas2017, Hoffmann2011, Andrist2015}. The interaction is long enough to expose participants to the full range of robot behaviours, solving it requires definitive interaction with the robot, and, at the same time, is not too cognitively demanding to distract from the robot.

Following the taxonomy outlined by Beer, et. al. \cite{Beer2014}, we have classified the robot as fully autonomous, drawing the action boundary at giving the user specific instructions what to do. While it was technically impossible for NAO to pick up the disks (they are too large) and modify the game by itself, NAO instructed participants to carry out its moves.

The technical setup is summarized in figure \ref{fig:technical_pipeline}. NAO used a video stream from its head camera, which we manually configured, to identify the game state. To ensure consistent recognition of the disks, we fixed the tower to the table and added a custom posture to NAOs posture library, to which it returned whenever it assessed the game state. We also added a light source behind NAO to ensure sufficient exposure and increase robustness towards changes in ambient light. To detect the presence of a disk on one of the poles, NAO used color thresholding of the disk's colors in three regions of interest, one for each pole\footnote{Source code available at: https://github.com/usr-lab/towerofhanoi }.

From here, NAO used breadth first tree search with pruning of previously visited states to compute the optimal sequence of moves, solving the game in the least number of turns. This method is run for both the robot's and the user's turns, allowing NAO to give feedback on the user's move and display appropriate behaviours.

\begin{table}[t]
    \centering
    \caption{Differences in the Robot's Social Behaviours}
    \label{tab:behaviours}
    \begin{tabular}{p{.25\linewidth}p{.3\linewidth}p{0.3\linewidth}}
    \toprule
    Behaviour & Social & Non-Social \\ \midrule
    Personalization & Uses 'we' & Uses third person \\
    \cellcolor{Gray}Feedback  & \cellcolor{Gray}Gives positive or negative feedback & \cellcolor{Gray}States if the action was optimal \\
    Verbal Phrases & Variety of phrases (3 or more variations) & Only one phrase per trigger \\
    \cellcolor{Gray}Gestures & \cellcolor{Gray}Pointing, Waving, Nodding, Thinking & \cellcolor{Gray}Pointing \\
    Gaze & Switches gaze between participant and game & Only looks at game \\
   \cellcolor{Gray} Memory & \cellcolor{Gray}Says that it really enjoys this game & \cellcolor{Gray}States that the user is the 27th person to play the game \\
    Feedback Frequency & Randomly with 33\% on optimal move, always on sub-optimal moves & For every sub-optimal move, and every 5th move. \\ \bottomrule
    \end{tabular}
\end{table}

To create the social behaviours, we reviewed the HRI literature identifying commonly used behaviours. Some of the most frequent examples include gaze cues, turn-taking, non-verbal gestures, personalisation, feedback, and memory \cite{Mutlu2006, Kose-Bagci2008, Breazeal2005, Corrigan2013, Ahmad2019, Fischer2013, Irfan2019}. Based on these, we developed one version of the robots behaviour depicting a 'social' robot, and one depicting a 'non-social' robot, see table \ref{tab:behaviours}. NAO launches these behaviours at specified trigger points based on comparisons between the observed and expected game state. We created a set of $6$ custom animations: wave, think (scratching it's head), nod, point left, point middle, and point right.

We then created a custom module for NAO which would run the game and added it to the naoqi framework via a custom broker. The module was run on a separate computer accessible via the network, and all the processing (CV, and AI) was done there.

\section{Experiment 1 (Framing Validation)}\label{sec:Experiment1}
\subsection{Hypothesis}

There will be a significant effect of framing, such that participants who read the high mind frame will have higher attributions of mind perception than those who read the low mind frame.

\subsection{Participants}\label{sec:experiment1_participants}
We recruited $110$ participants from the online platform Amazon Mechanical Turk (AMT). To ensure sufficient English proficiency, we only recruited participants from countries with English as the official language (US, Canada, UK, Australia, New Zealand). Further, to ensure high response quality, we limited participation to workers with a $99\%$ approval rating and introduced attention checks prior to and during the experiment.

Out of the $110$ participants, $33$ were discarded due to failing attention checks. The remaining $77$ participants ($M_{age}=28.05$, $SD=12.16$) were randomly assigned into two conditions: high-mind ($M_{age}=27.79$, $SD=12.28$), and low-mind ($M_{age}=28.31$, $SD=12.20$). There were no significant differences in age ($F(1,76)=.034$, $p=.854$) or gender ($X^2(1, N=77)=0.21$, $p=.883$) between the groups.

The survey took approximately 5 minutes to complete, and participants were compensated $0.8$ USD for their time.

\subsection{Material}
To control for participant's conceptions of robots when filling out the questionnaire, we showed them a picture of a NAO robot\footnote{{https://www.softbankrobotics.com/emea/en/nao}}. This image was then combined with a text description about the robot's capabilities. 

To measure mind perception, we used the dimensions of mind perception questionnaire introduced by \citet{Gray2007}. However, we modified the initial scale by replacing the original 7-point Likert scale with semantic anchors with a 5-point Likert scale ranging from \textit{not capable at all} to \textit{extremely capable}. This allowed us to investigate mind perception for individual robots, rather than having to compare two robots. 

The image, description, and questionnaire were shown on a single page and provided via the participant's browser using a survey constructed with UniPark survey software. 

\subsection{Design and Procedure}\label{sec:Experiment1_DesignAndProcedure}
We employed a 1-way independent groups design. Ethics approval was obtained from the Bremen International Graduate School of Social Sciences (BIGSSS). Participants were first shown an information sheet which detailed the procedure and informed them of their right to withdraw at any time, after which they indicated their consent to participate. Next, participants were shown a picture of a NAO robot and asked four attention check questions about the image. If any of these questions were answered incorrectly, the survey would end without presenting the manipulation or measurement to the participant. If participants answered all attention check questions correctly, they were randomly assigned to one of the two levels of framing. 

Participants were then presented with the same picture of the robot, but now accompanied with one of the two frames (high/low mind) manipulated as the independent variable\footnote{The exact manipulations are available in the supplementary material.}. We designed both descriptions to convey the same factual information. The dependent variable was the modified version of the mind perception questionnaire \cite{Gray2007}. Participants were instructed to look at the picture, read the text below, and then fill out the mind perception questionnaire. The order of scale items was randomized for each participant.

Next, we added a final attention check, asking about the role of the robot as described by the frame (both frames had the same role of the robot as a teacher). This was done to ensure that participants had thoroughly read the description. Afterwards, participants were asked to provide demographics. 

Finally, participants were taken to a debriefing statement informing them about the two conditions of the study, the aim, and the contact details of the experimenters should they have any questions about the study. 

\subsection{Results and Discussion}
After collecting the data, reliability of the overall scale was computed. We excluded participants that were missing more than 20\% of their data. The reliability of the overall $16$ item scale, \textit{mind perception}, was $\alpha = 0.94$ ($N=64$ valid cases), indicating the questionnaire is highly reliable.

\begin{table}[t]
    \centering
    \caption{Means and SDs Experiment 1}
    \label{tab:experimental_means_framing}
    \begin{tabular}{@{}lllllll@{}} 
    \toprule
                & Condition       & Mean & SD & N & M & F   \\ \midrule 
mind perception & low-mind & 2.14 & 0.8 & 39 &24&14 \\ 
                & high-mind & 2.79 & 0.91 & 39 &24&15 \\ 
                & total & 2.46 & .91 & 78 &48&29\\\bottomrule 
    \end{tabular} 
\end{table}

We then conducted an independent samples t-test to compare mind perception in the high and low mind framing conditions. The mean and standard deviations for each group are reported in table \ref{tab:experimental_means_framing}. There was a significant difference in mind perception between participants who viewed the high-mind frame, and those who viewed the low-mind frame, $t(76)=3.369$, $p<0.001$, $d=0.763$ (see figure \ref{fig:experiment1_results}). This indicates the framing manipulation was successful in influencing participants mind perception attributed to the robot, with a moderate to high effect size. 

\section{Experiment 2 (Behaviour Validation)}\label{sec:experiment2}
\subsection{Hypothesis}
There will be a significant effect of behaviour, such that participants who see the video of the social robot will have higher attributions of mind perception than those who see the video of the non-social robot.

\subsection{Participants}
Participants were again recruited from AMT, with the same participation criteria as in Experiment 1 (see Section \ref{sec:experiment1_participants}). $232$ participants completed the survey, of which $170$ failed one of the attention checks, leaving $62$ ($M_{age} = 38.72$, $SD = 11.85$) eligible participants. These participants were then randomly assigned to one of the two conditions; $33$ viewed the social robot videos ($M_{age}=40.27,~SD=12.04$), and $28$ viewed the non-social videos ($M_{age}=36.89,~SD=11.56$). There was no difference between gender ($X^2(1, N=61)=.350,~p=.554$) or age ($F(1,59)=1.239,~p=.270$) across conditions.

The survey took approximately $5$ minutes to complete, and participants were compensated $0.8$ USD for their time.

\subsection{Material}
For this experiment, we used videos showing NAO introducing the Tower of Hanoi, and then playing it with the experimenter. Each condition (social/non-social) had two videos. The first video of the set showed NAO introducing itself and explaining the rules of the game. The second video showed three scenes from a game played between NAO and the experimenter\footnote{Videos are available in the supplementary material.}. The videos were filmed from the perspective a participant may have during the real-world experiment, and the scenes were designed to show all the differences present in the two conditions. The experimenter was not seen, aside from their hand.

Mind perception was again measured using the modified version of the mind perception questionnaire by \citet{Gray2007}. We used the Limesurvey survey tool to host the survey.

\subsection{Design and Procedure}
Ethics approval for this study was obtained from the Jacobs University Ethics Committee. We used a similar design as in our first experiment, but replaced the initial image participants saw when answering the attention check questions with the first video from the set of their assigned condition (duration 40s). We also replaced the following image and description of the robot with the second video of the set (duration $160 s$).

Upon entering the survey, participants were randomly assigned to either the social or non social behaviour condition as the independent variable. From then on, the procedure was the same as in Experiment 1, see section \ref{sec:Experiment1_DesignAndProcedure}.

To ensure that we did not induce any false expectations about the robot during the experiment, the debriefing statement contained an additional paragraph indicating that the robot's agency may appear different from its true capabilities. We also included a link to Softbank Robotics's website, should participants wish to obtain more information.  

\subsection{Results and Discussion}
Reliability of the mind perception score ($18$ items, $\alpha=.92$) was high. We again excluded participants if more than $20\%$ of the responses were missing.

\begin{figure}[t]
\centering
\begin{subfigure}{.45\linewidth}
\includegraphics[width=\linewidth]{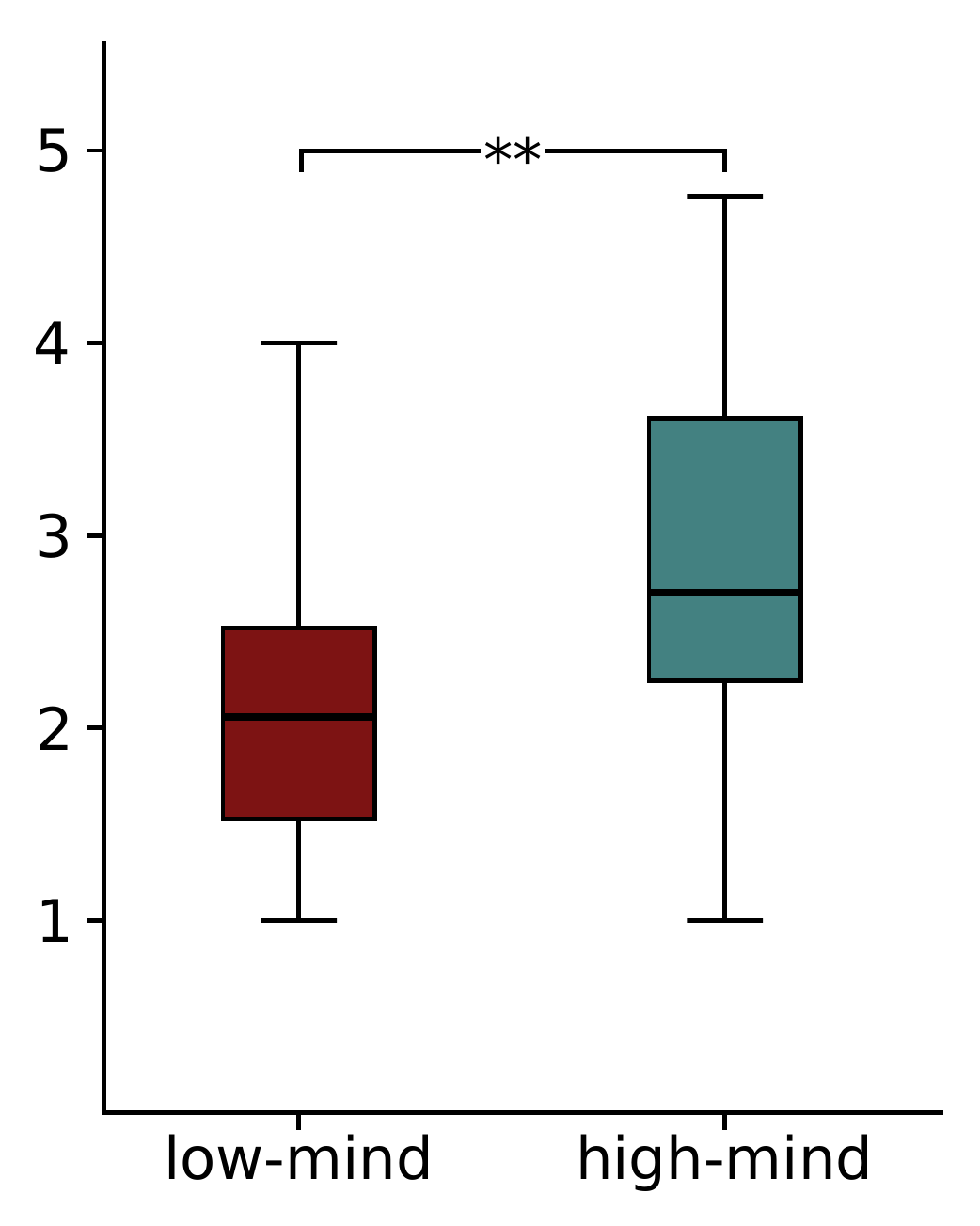}
\caption{framing}
\label{fig:experiment1_results}
\end{subfigure}
\begin{subfigure}{.45\linewidth}
\includegraphics[width=\linewidth]{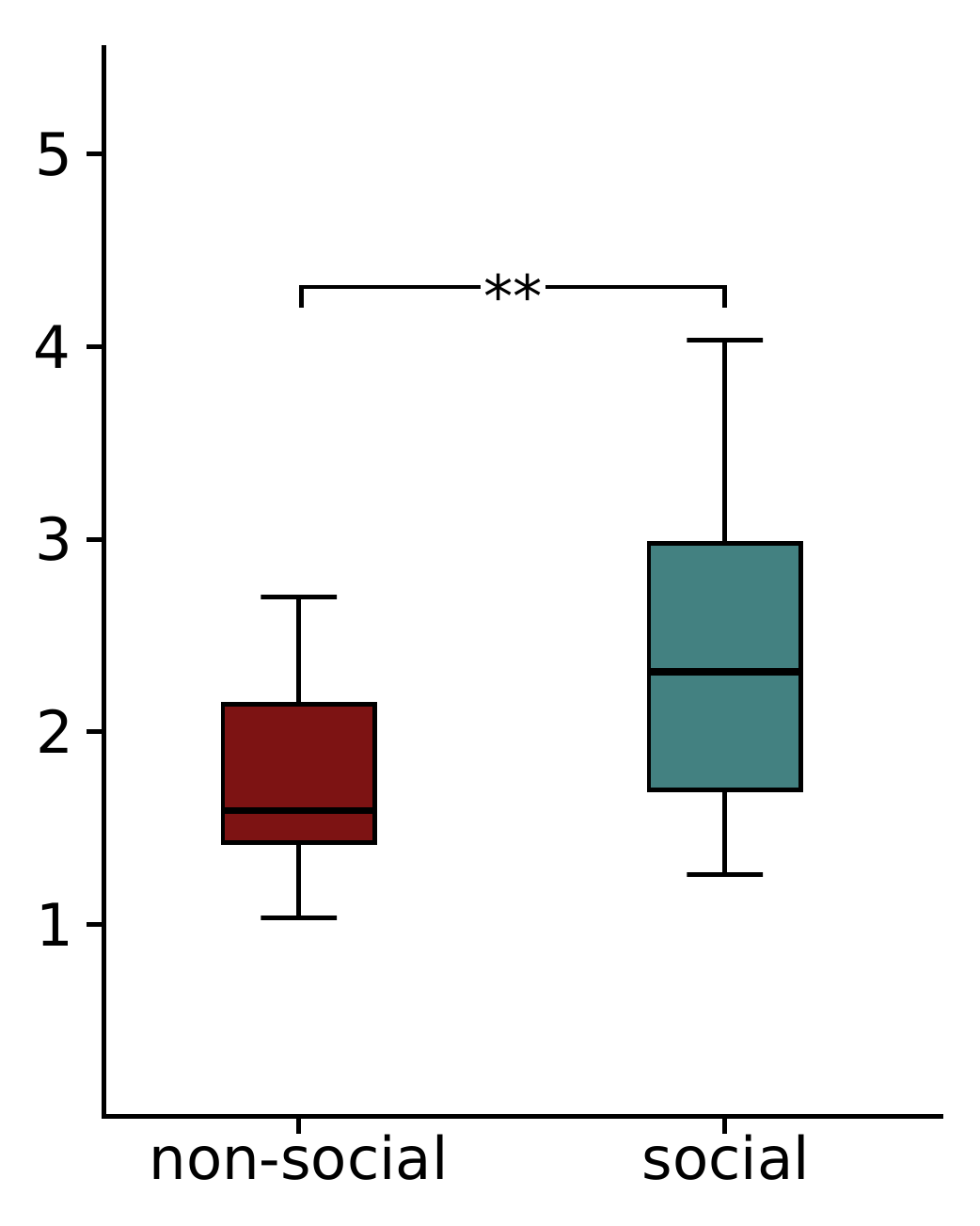}
\caption{behaviour}
\label{fig:experiment2_results}
\end{subfigure}
\caption{The distribution of mind perception scores in experiment 1 and experiment 2. Mind perception differs significantly ($p<.01$) between levels for both framing (left) and behaviour (right).}
\end{figure}

We performed an independent samples t-test on mind perception  ($t(60)=3.2$, $p=.002$, $d=.82$), comparing the social and non-social behaviours; the results are shown in figure \ref{fig:experiment2_results}. This suggests the social behaviours were successful in increasing participants attributions of mind perception, with a large effect size. 

\begin{table}[t]
\centering
\caption{Results Experiment 2}
\label{tab:experiment2-results}
\begin{tabular}{@{}lllllll@{}}
\toprule
               & behaviour   & Mean    & SD  & N & M & F \\ \midrule
MindPerception & non-social & 1.79 & .68 & 28 &14&14                    \\
               & social     & 2.39 & .76 & 34 &14&19                   \\
               & total      & 2.12 & .78 & 62 &28&33                   \\ \bottomrule
\end{tabular}
\end{table}

\section{Experiment 3 (Interaction)}
\subsection{Hypotheses}
(H1) There will be a significant effect of framing, such that participants who read the high mind frame will have higher attributions of mind perception and moral concern than those who read the low mind frame.
(H2) There will be a significant effect of behaviour, such that participants who interact with the social robot will have higher attributions of mind perception and moral concern than those who interact with the non-social robot.
(H3) There will be a significant framing by social behaviours interaction, where behaviour will have a stronger effect than framing on mind perception and moral concern. That is, the difference between social and non-social behaviours for the high mind frame will be lower than the difference between social and non-social behaviours for the low mind frame.

\subsection{Participants}
We recruited students from Uppsala University via advertisements on notice boards across the entire university, with particular focus on the buildings for the humanities, medicine, and psychology. This was done to attract more people with a non-technical background. We also advertised the experiment in lectures, and recruited participants directly by approaching them and inviting them to participate.

$100$ participants ($M_{age}=24.36$, $SD=4.55$) completed the experiment. Participants were randomly assigned to one of $4$ conditions: (1) low-mind, non-social ($M_{age}=24.84,~SD=5.67$), (2) low-mind, social ($M_{age}=24.67,~SD=4.46$), (3) high-mind, non-social ($M_{age}=24.38,~SD=3.94$), (4) high-mind, social ($M_{age}=23.84,~SD=4.07$). $1$ participant had to be excluded due to technical difficulties with the robot (faulty CV pipeline), leaving $99$ eligible participants. Each condition had $25$ participants, with exception of the high-mind, non-social condition, which had $24$.

The experiment took approximately $25$ minutes to complete, and participants were compensated with a voucher worth approximately $5$ USD for their time.

\subsection{Design}
Ethics approval was obtained from the Jacobs University Ethics Committee. We employed a 2-way full factorial, independent groups design. Our independent variables were framing (high/low mind, see section \ref{sec:Experiment1}), and robot behaviour (social/non-social, see section \ref{sec:experiment2}). The number of participants needed to detect an effect with 95\% power and $\alpha = 0.05$, as recommended by \citet{Cohen1992}, was calculated \textit{a priori} using G*Power Software \cite{Faul2007}. Although moderate-large effect sizes were found in Experiments 1 and 2, we chose to be conservative with a moderate effect size estimation ($f^2$ = 0.15; \cite{Cohen1992}). Following these par ameters, the number of participants required to detect a medium effect size is \textit{N} = 73. Given our sample size of $N=99$, our analyses should be sufficiently powered to detect a medium to large effect. 

Our dependent variables were mind perception \cite{Gray2007}, moral concern \cite{Nomura2019}, as well the Godspeed questionnaire \cite{Bartneck2009a}. Participants also answered 3 questions regarding their familiarity with robots in general, familiarity with the NAO robot specifically, and openness to new technologies. 

\subsection{Material}
For this experiment, participants interacted with the NAO robot directly, using the Tower of Hanoi as an interaction task. NAO was programmed as described in section \ref{sec:technical_spec}. Additionally, we provided a laptop that participants used to fill in the questionnaires before and after the interaction. If additional consent was given, the entire experiment was video recorded from two angles, a front angle, and a side angle.

To measure mind perception, we again used the adapted mind perception questionnaire by \citet{Gray2007}.
To measure the amount of moral concern participants felt towards NAO, we used a modified version of the Measurement of Moral Concern for Robots scale \cite{Nomura2019}. We reformulated each item to start with the phrase "I would", and reverse coded the item, if the original started with "I wouldn't". The final scale had 30 items, out of which 12 were reverse coded.

Finally, all dimensions of the Godspeed questionnaire \cite{Bartneck2009} were given to participants to measure anthropomorphism of the robot.

\subsection{Procedure}
After entering the lab, participants were asked to sit in front of the Tower of Hanoi opposite of NAO. Then, participants were asked to read and sign the prepared information sheet and consent form. This included the optional decision to have the interaction video recorded for later analysis. Afterwards, participants were directed to a laptop, next to the tower, where they filled in a demographic questionnaire and answered questions about their familiarity with robots, NAO, and openness to new technology. Following this, they were presented with the framed description of the robot (high/low mind). This description was shown automatically after the demographics questionnaire was completed. After reading the description, participants were prompted to stop filling out the questionnaire and asked to inform the experimenter, so that they could start the game. Participants then played the game with NAO showing one of the two sets of behaviour (non-social/social). Once the game was completed, participants were asked to continue filling out the questionnaire, which presented the three measures in random order. Questions within each measure were also randomized. After completing the survey, participants were shown a debriefing statement, thanked for their participation in the study, and given the voucher as compensation for their time.

\subsection{Results and Discussion}

\begin{figure*}[t]
    \centering
    \begin{subfigure}{.25\linewidth}
    \includegraphics[width=\linewidth]{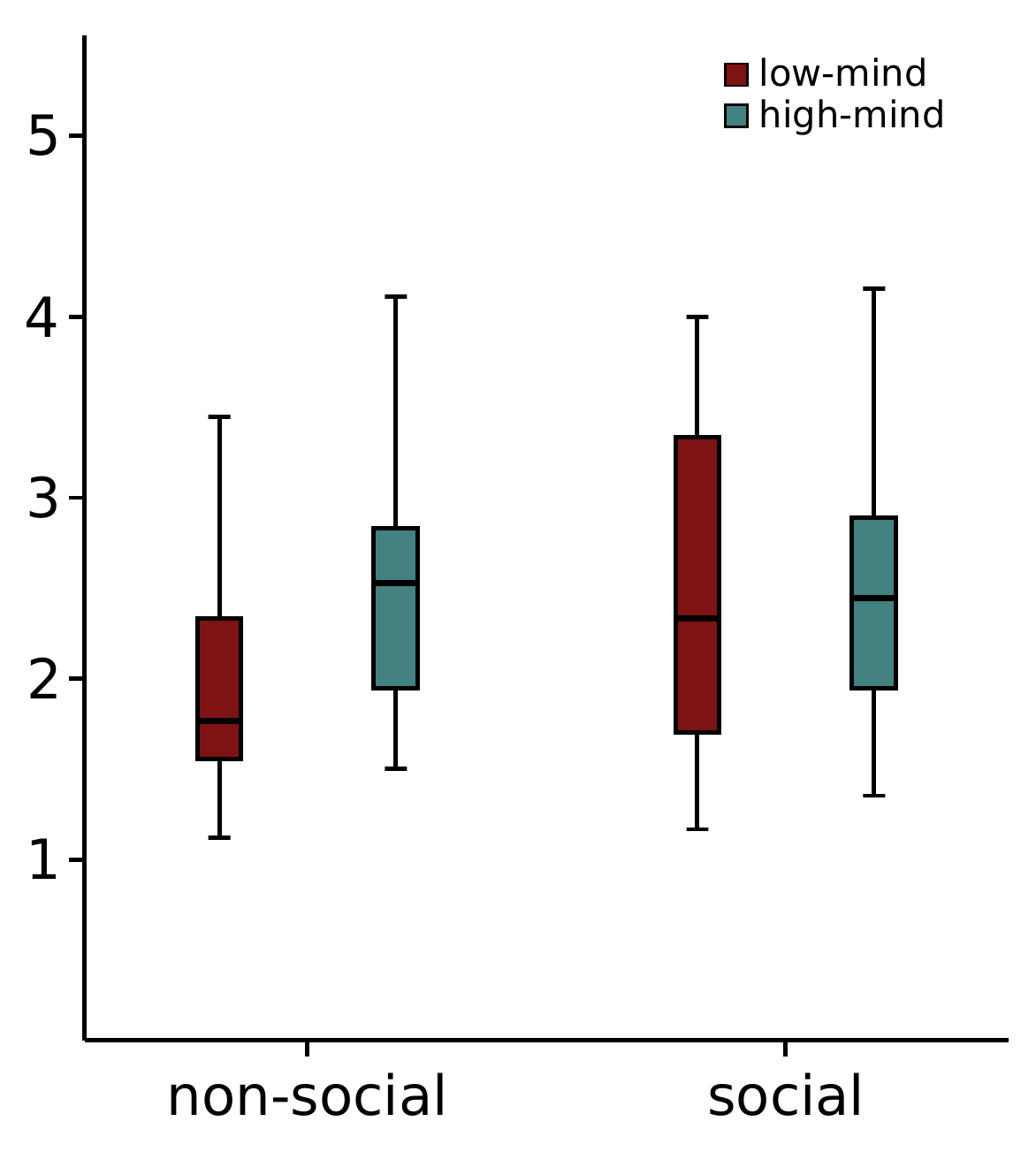}
    \caption{Mind Perception}
    \label{fig:experiment3_mind_perception}
    \end{subfigure}
    \begin{subfigure}{.25\linewidth}
    \includegraphics[width=\linewidth]{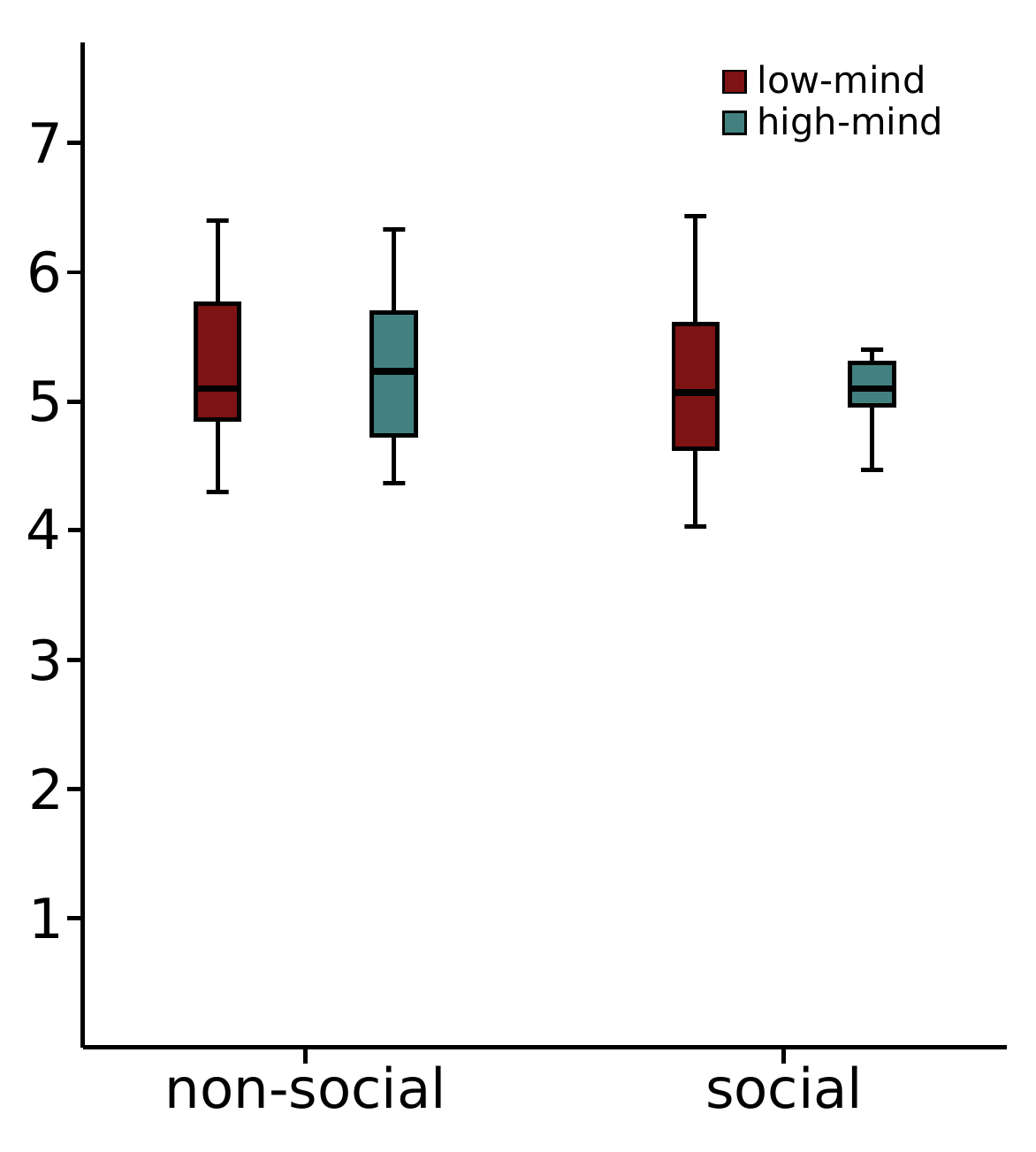}
    \caption{Moral Concern}
    \label{fig:experiment3_moral_concern}
    \end{subfigure}
    \begin{subfigure}{.25\linewidth}
    \includegraphics[width=\linewidth]{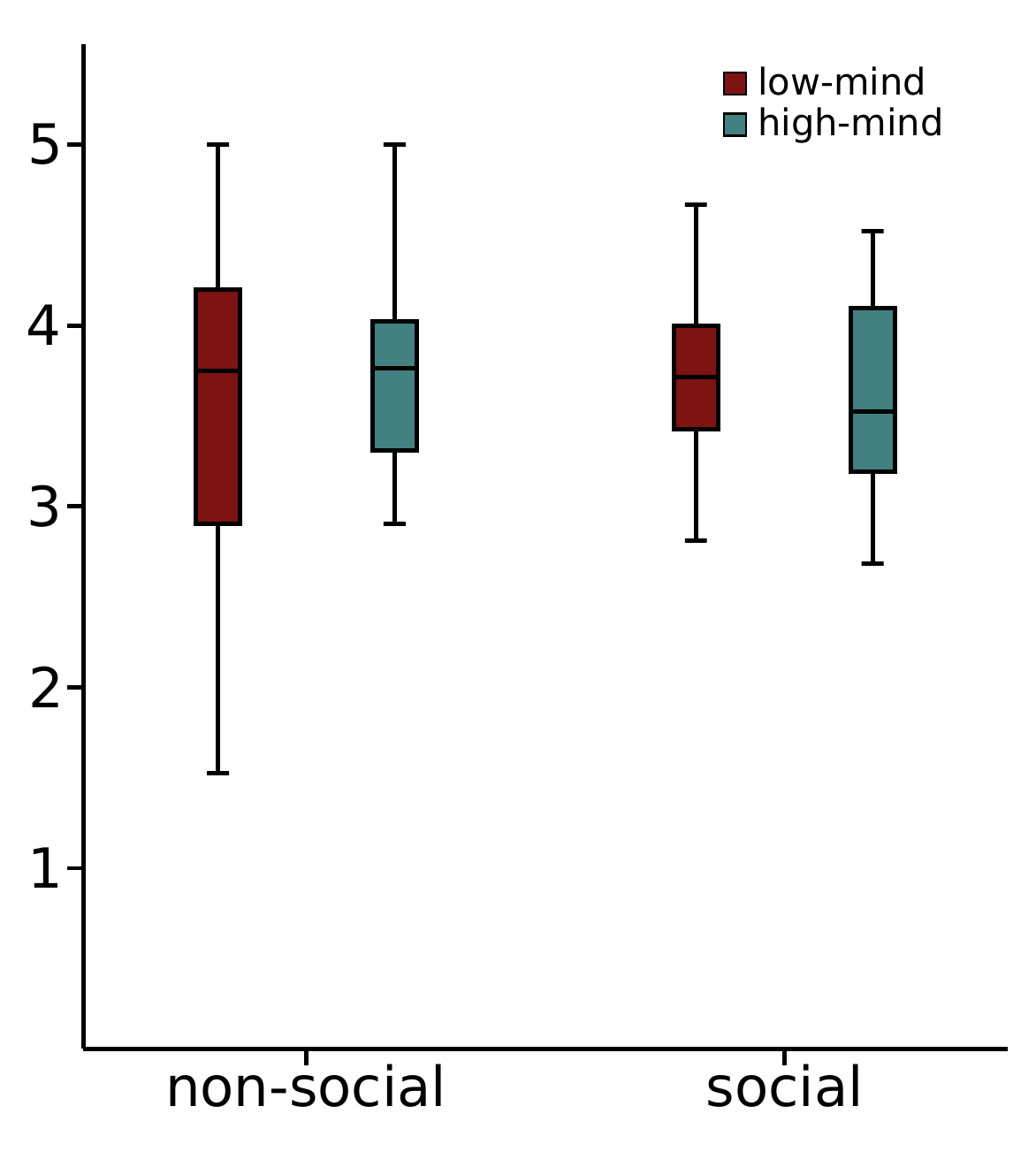}
    \caption{Godspeed}
    \label{fig:experiment3_godspeed}
    \end{subfigure}
    \caption{The main result of experiment 3. None of the measures differ significantly between either framing, or behaviour. While there is no significant interaction, a trend is visible in the data for the non-social condition and different frames on mind perception.}
    \label{fig:experiment3_results}
\end{figure*}

Before analyzing the effect of our conditions, we verified the reliability of the measures. Mind Perception ($N=18$, $\alpha = .92$), Moral Concern ($N=30$, $\alpha = .87$) and Godspeed ($N=21$, $\alpha=.95$) were all found to be highly reliable.
The reliability of the perceived safety scale of the Godspeed questionnaire was low ($N=3$, $\alpha = .53$). We therefore excluded this sub-scale from our final analyses.

We then proceeded to check for an effect of three potential covariates: age, gender, and technical background. The latter was measured using the three questions outlined in the procedure's pre-test, which we tested independently. First, we checked for correlation between the potential covariates and the outcome measures. The only significant correlation was between gender and moral concern ($\sigma=-.299$, $p<0.03$), indicating that women showed greater moral concern for the robot than men.

During this analysis we also noted significant correlations between the dependent variables. There was a significant correlation between mind perception and moral concern ($\sigma=.405$, $p<0.001$), mind perception and Godspeed ($\sigma=.527$, $p<.001$), and Godspeed and moral concern ($\sigma=.601$, $p<.001$). These findings support our hypotheses that these three constructs are related, and further suggests using a MANOVA to account for dependencies between these variables rather than separate ANOVAS.

Afterwards, we tested for an effect of gender between conditions using a $X^2$-test; we found no significant difference ($X^2(3, N=99)=1.356$, $p>>.05$), which is unsurprising given our random assignment of participants to conditions. As correlation is a necessary condition for determining confounds, mediation, or moderation, we concluded that none of the measured potential confounds affected our experiment.

\begin{table}
\caption{Results Experiment 3}
\label{tab:experiment3-means}
\begin{tabular}{@{}p{0.15\linewidth}lllllll@{}}
\toprule
               & behavior   & framing     & Mean    & SD  & N & M & F \\

\midrule
\multirow{2}{0.15\linewidth}{Mind Perception} & non-social & low-mind & 2.15 & .89 & 25 &17&8                  \\
               &            & high-mind & 2.48 & .64 & 24 &16&8                   \\
               &            & total       & 2.31 & .79 & 49 &33&16                   \\\cmidrule{2-8}
               & social     & low-mind & 2.47 & .86 & 25  &14&11                  \\
               &            & high-mind & 2.54 & .73 & 25 &14&11                   \\
               &            & total       & 2.50 & .79 & 50 &28&22                   \\\cmidrule{2-8}
               & total      & low-mind & 2.31 & .88 & 50  &31&19                  \\
               &            & high-mind & 2.50 & .68 & 49 &30&19                   \\
               &            & total       & 2.40 & .79 & 99 &61&38                   \\\midrule
\multirow{2}{0.15\linewidth}{Moral Concern}   & non-social & low-mind & 5.17 & .84 & 25  &17&8\\
               &            & high-mind & 5.18 & .79 & 24 &16&8                   \\
               &            & total       & 5.17 & .81 & 49 &33&16                   \\\cmidrule{2-8}
               & social     & low-mind & 5.07 & .84 & 25  &14&11                  \\
               &            & high-mind & 5.09 & .56 & 25 &14&11                   \\
               &            & total       & 5.08 & .60 & 50 &28&22                   \\\cmidrule{2-8}
               & total      & low-mind & 5.12 & .74 & 49  &31&19                  \\
               &            & high-mind & 5.13 & .68 & 50 &30&19                   \\
               &            & total       & 5.13 & .71 & 99 &61&38                   \\\midrule
Godspeed       & non-social & low-mind & 3.53 & .85 & 25  &17&8                  \\
               &            & high-mind & 3.8  & .57 & 24 &16&8                   \\
               &            & total       & 3.67 & .73 & 49 &33&16                   \\\cmidrule{2-8}
               & social     & low-mind & 3.63 & .61 & 25  &14&11                  \\
               &            & high-mind & 3.6  & .56 & 25 &14&11                   \\
               &            & total       & 3.61 & .58 & 50 &28&22                   \\\cmidrule{2-8}
               & total      & low-mind & 3.58 & .73 & 49  &31&19                  \\
               &            & high-mind & 3.70 & .57 & 50 &30&19                   \\
               &            & total       & 3.64 & .66 & 99 &61&38                  \\\bottomrule
\end{tabular}
\end{table}

Following this, we ran a MANOVA using framing and behaviour as independent variables, and our three measures (Mind Perception, Moral Concern, Godspeed) as dependent variables. Contrary to our hypotheses, we found no significant effect of either (H1) framing ($F(3, 93)=.735$, $p=.534$, Wilkin's $\Lambda=.977$), nor  (H2) behaviour ($F(3, 93)=1.123$, $p=.344$, Wilkin's $\Lambda=.965$). The (H3) interaction between framing and behaviour was also non-significant ($F(3, 93)=.704$, $p=.552$, Wilkin's $\Lambda=.978$). The group means are denoted in table \ref{tab:experiment3-means} and visualisations for the distribution for mind perception and moral concern can be found in figure \ref{fig:experiment3_results}.

\section{General Discussion}
The most unexpected finding of our research is that the manipulations tested in the online experiments did not replicate in the real world. The non-significant findings on any of the three scales (mind perception, moral concern, and anthropomorphism) also means that we were unable to further analyse the relationship between these constructs. The lack of replication also suggests that there is at least one other factor at play that is confounding the manipulations. There are two main potential sources of difference in this experiment: population and experimental setting. As the population factors age, gender, and technical background were assessed as potential confounds during the experiment, we can rule these out as causes; this makes the experimental setting - online vs. real-world - the more likely cause.

The first main source of difference between the three experiments is the population tested. The population in the first two experiments was recruited from AMT; the population in the third experiment consisted of European university students. As stated above, we can rule out age, gender and technological background, due to non-significant correlation with either the independent or dependent variables. The exception here is moral concern, which differs significantly by gender, but doesn't differ significantly between conditions. Another possible confound between the two could be cultural background; however, this seems unlikely due to the large diversity of both the AMT population and the university population. In addition, the original mind perception sub-scales have been replicated cross culturally in a Japanese sample, suggesting that mind perception is a culturally generalizable phenomenon \cite{Ishii2019}. As such, we think that it is unlikely - though not impossible - that differences in population are causing the difference in effect size between experiments.

The second source of difference between the experiments comes from the environment. For the first two experiments the robot was embodied virtually (picture and video), whereas the final experiment involved physical embodiment. Additionally, the role of participants differed between experiments. Whilst the first two experiments were online, and therefore involved the participants only as observers, in the final experiment they were able to interact with the robot directly. The duration of the experiment was also extended from around 5 minutes in the first two experiments to 25 minutes in the third. The third experiment was performed in a research lab, whereas the first two were carried out at a location of the participant's choosing. 

One consequence of moving from a virtual embodiment with participants as observers to a real-world interaction with a physically embodied robot could be change in participant's attention. While the focus of experiment 1 and experiment 2 was clearly on the robot, participants in experiment 3 had to split their attention between the robot and the game. Potentially, the focus on the task may have distracted from the robot's behaviours, blurring out the differences between conditions due to lower engagement with the robot. However, in this case we would still expect a significant effect of framing - as frames were presented before the game could become a distractor -, and we would expect that participants that engage more with the robot show a larger effect on the measures. We can use time spent playing the game as a proxy measure for engagement. However, we do not find any significant correlation ($\sigma=.114,~p=.236$) between completion time and participant's ratings on mind perception, making lack of focus on the robot an unlikely explanation.

A second possibility is that real-world interaction is a lot richer than merely seeing a picture or video. Participants may have more sources from which they draw when assessing the extent of the robot's mind in the real-world interaction, making effects harder to isolate. This would also explain the lack of replication for the framing manipulation, as other factors present during the interaction could outweigh the effect of the frame. In addition, factors like in-group affiliation, which aren't present in virtual scenarios due to the nature of the participant being an observer, may be present in real-world interactions. More factors at play would mean that the effect size could be lower in the real-world. If we add the type of experiment (virtual / real-world) as another variable, we can test the assumption that real-world interaction leads to lower effect size. This decrease would be visible as an interaction between type of experiment and manipulation. That is, we would expect the effect of framing and behaviour on mind perception to be significantly less in the real-world experiment than in the two pilots, where only an image and video were used. Hence, we combined our data into one large dataset, possible due to the non-significant interaction in experiment 3, and performed two 2-way ANOVAs comparing both framing and behaviour across the type of experiment (online with picture/video versus real-world interaction).

\begin{figure}
    \centering
    \begin{subfigure}{.45\linewidth}
        \includegraphics[width=\linewidth]{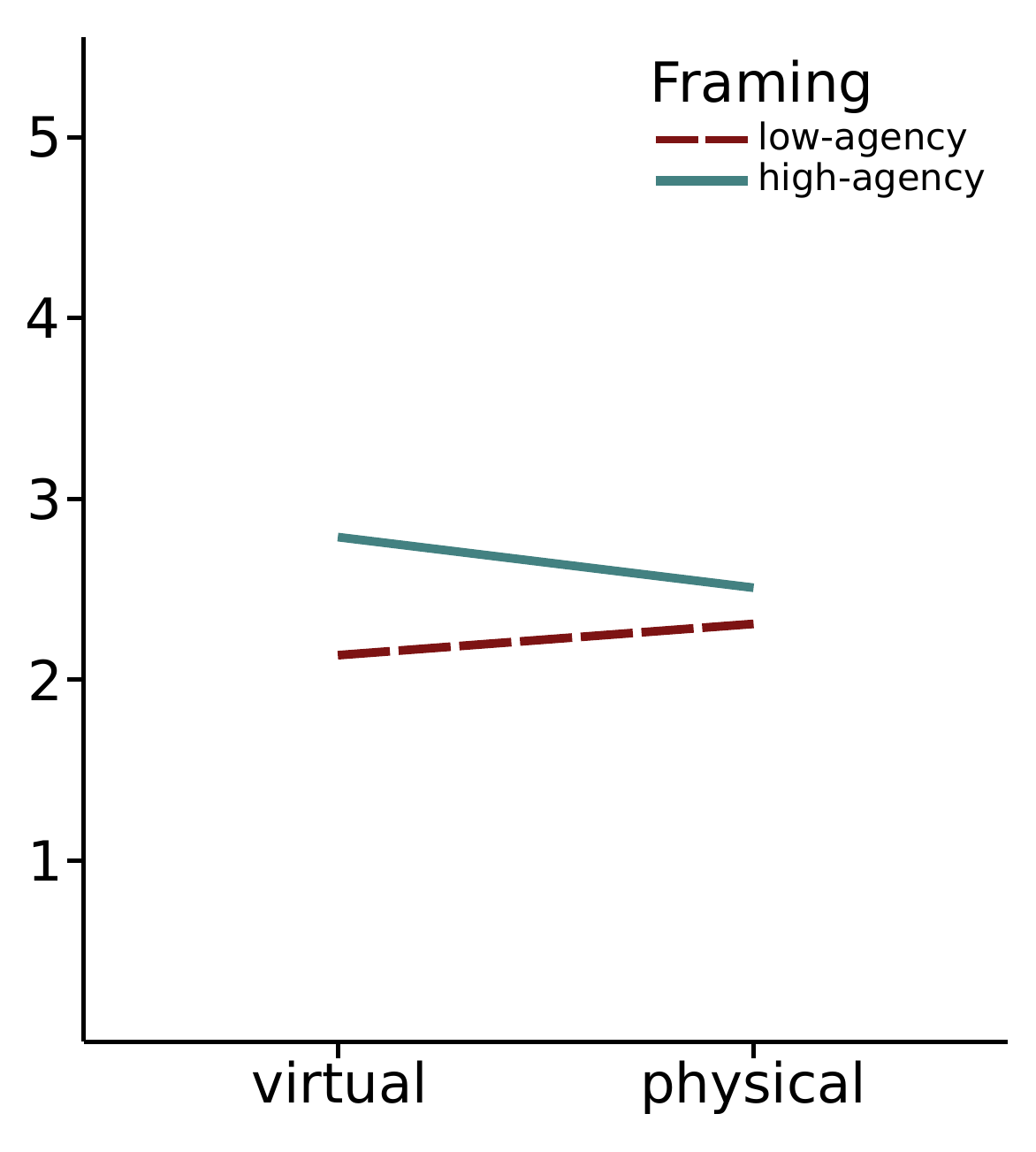}
        \caption{framing}
    \end{subfigure}
    \begin{subfigure}{.45\linewidth}
        \includegraphics[width=\linewidth]{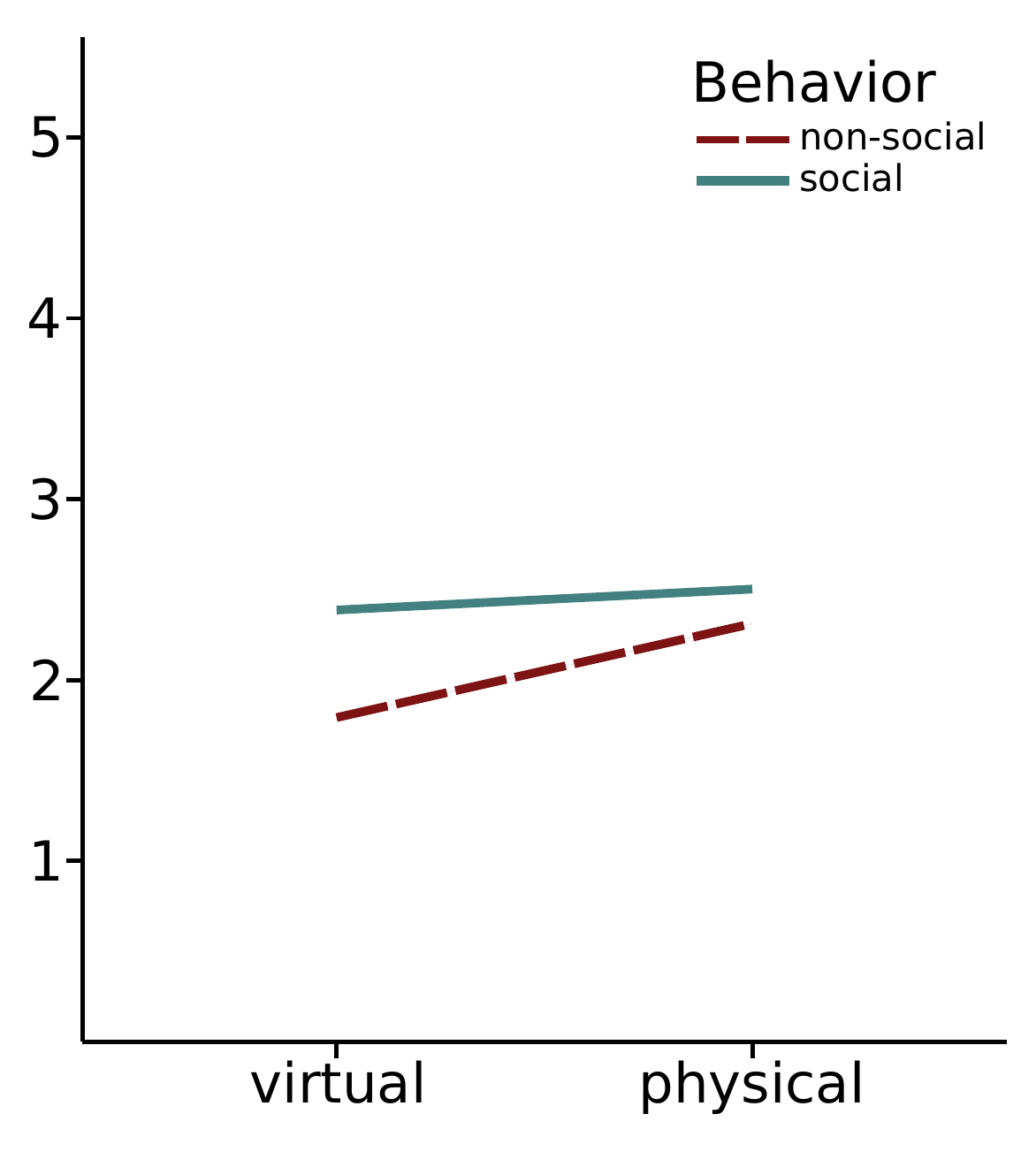}
        \caption{behaviour}
    \end{subfigure}
    \caption{A contrast between the virtual experiment and the real-world interaction on the mean score of mind perception split by manipulations. We can see a strong indication for an interaction between experiment type and manipulation.}
    \label{fig:post-hoc-anova}
\end{figure}

First, we conducted an ANOVA testing the interaction between the type of experiment (virtual / physical) and framing (high/low mind), ($F(3,173)=4.691,~p=.004$) see figure \ref{fig:post-hoc-anova}. We found a non-significant interaction between the experimental conditions and framing ($p=.069$), a significant main effect of framing ($p=.001$), likely driven by the effect of Experiment 1, and a non-significant main effect of the experimental condition ($p=.667$). However, if the effect size in the real-world interaction is indeed small, more power may be required (suggested $N > 200$), making this (post-hoc) analysis difficult to interpret. Consequently, we recommend a follow-up study with higher power more specifically targeted towards comparing the virtual and real-world setting in the context of framing. 

Next, we conducted a second ANOVA  comparing the type of experiment (virtual / physical) and behaviour (social / non-social), ($F(3, 157)=5.421,~p=.001$), see figure \ref{fig:post-hoc-anova}. The ANOVA showed a non-significant interaction between experimental conditions and behaviour ($p=.107$), a significant main effect of behaviour ($p=.002$), again probably driven by Experiment 2, and a significant main effect of the experimental condition ($p=.012$). The main effect of the type of experiment could suggest that ratings of mind perception are higher for physically present robots than robots depicted in videos; again, however, a more rigorous follow-up experiment would be needed to determine this with certainty. 

Hence, a tentative explanation for what caused the reduction in effect size is that there may be more factors at play (such as intergroup dynamics) in the real-world setting, than were in either online experiment. This explanation also aligns nicely with previous work comparing virtual and physical embodiment. While some studies suggest that physically embodied robots lead to higher social presence \cite{Jung2013}, elicit higher ratings of empathy \cite{Seo2015}, and lead to greater engagement and enjoyment of the interaction \cite{Deng2019}, other studies contradict these findings \cite{Schneider2017, Ligthart2015}. In our experiment, we can equally see that physical embodiment leads to a consistent, high attribution of mind across conditions; however, the effect size of framing and behaviour is larger in the virtual pilots.

An appealing explanation for this is that \textit{human's perception of other minds is the result of interference between different sources of truth}. In a physical interaction these factors are, as mentioned above, harder to isolate. As we can only measure the resulting, inferred mean, the effect size of an individual factor will be reduced. In the virtual setting, however, the manipulation is more isolated, meaning less factors contribute overall, and variance in the manipulation is easier to detect. 

Should interference from other, non-measured factors be the main cause of our findings, this could also explain some of the other contradictory findings in research on physical and virtual embodiment. However, and this is a clear limitation of our work, we can only tentatively suggest this explanation, as we set out to test a different hypothesis and derived this explanation post-mortem. Hence, we highly encourage more research testing both, which factors contribute to outcomes like mind perception, and what differences exist between virtual and physical interactions. Ideally, specific factors which could potentially contribute to mind-perception in-vivo would need to be identified and manipulated individually. Additionally, we think that a meta-analysis on the effect of robot behaviours could help investigate if our theory can explain some of the conflicting findings between previous studies.

\section{Conclusion}
In summary, this paper shows evidence that mind perception is harder to manipulate in physical experiments than in virtual ones. Both experiment 1 and experiment 2 showed significant effects of framing and social behaviour on mind perception, respectively. However, these effects failed to replicate in a real-world setting. We tentatively suggest that this is caused by virtual interactions being more isolated, i.e., only providing a slice of the real interaction. We hypothesize that this could explain some of the contradictory findings in experiments between virtual and physical embodiment, although further research is needed to claim this with certainty.

\begin{acks}
Special thanks to Tatsuya Nomura for providing a translated version of the Measurement of Moral Concern for Robots scale, and to the ANIMATAS project's independent ethics advisor Dr. Agn\`es Roby-Brami, for providing additional thoughts on the experimental design.

This project has received funding from the European Union's Horizon 2020 research and innovation programme under grant agreement No 765955.
\end{acks}

\bibliographystyle{ACM-Reference-Format}
\bibliography{Paper_Refs.bib}


\begin{thebibliography}{57}


\ifx \showCODEN    \undefined \def \showCODEN     #1{\unskip}     \fi
\ifx \showDOI      \undefined \def \showDOI       #1{#1}\fi
\ifx \showISBNx    \undefined \def \showISBNx     #1{\unskip}     \fi
\ifx \showISBNxiii \undefined \def \showISBNxiii  #1{\unskip}     \fi
\ifx \showISSN     \undefined \def \showISSN      #1{\unskip}     \fi
\ifx \showLCCN     \undefined \def \showLCCN      #1{\unskip}     \fi
\ifx \shownote     \undefined \def \shownote      #1{#1}          \fi
\ifx \showarticletitle \undefined \def \showarticletitle #1{#1}   \fi
\ifx \showURL      \undefined \def \showURL       {\relax}        \fi
\providecommand\bibfield[2]{#2}
\providecommand\bibinfo[2]{#2}
\providecommand\natexlab[1]{#1}
\providecommand\showeprint[2][]{arXiv:#2}

\bibitem[\protect\citeauthoryear{Abubshait and Wiese}{Abubshait and
  Wiese}{2017}]%
        {Abubshait2017}
\bibfield{author}{\bibinfo{person}{Abdulaziz Abubshait} {and}
  \bibinfo{person}{Eva Wiese}.} \bibinfo{year}{2017}\natexlab{}.
\newblock \showarticletitle{{You Look Human, But Act Like a Machine: Agent
  Appearance and Behavior Modulate Different Aspects of Human-Robot
  Interaction}}.
\newblock \bibinfo{journal}{\emph{Frontiers in psychology}}
  \bibinfo{volume}{8}, \bibinfo{number}{August} (\bibinfo{year}{2017}),
  \bibinfo{pages}{1--12}.
\newblock
\urldef\tempurl%
\url{https://doi.org/10.3389/fpsyg.2017.01393}
\showDOI{\tempurl}


\bibitem[\protect\citeauthoryear{Ahmad, Mubin, Shahid, and Orlando}{Ahmad
  et~al\mbox{.}}{2019}]%
        {Ahmad2019}
\bibfield{author}{\bibinfo{person}{Muneeb~Imtiaz Ahmad}, \bibinfo{person}{Omar
  Mubin}, \bibinfo{person}{Suleman Shahid}, {and} \bibinfo{person}{Joanne
  Orlando}.} \bibinfo{year}{2019}\natexlab{}.
\newblock \showarticletitle{{Robot's adaptive emotional feedback sustains
  children's social engagement and promotes their vocabulary learning: a
  long-term child-robot interaction study}}.
\newblock \bibinfo{journal}{\emph{Adaptive Behavior}} \bibinfo{volume}{27},
  \bibinfo{number}{4} (\bibinfo{year}{2019}), \bibinfo{pages}{243--266}.
\newblock
\showISSN{17412633}
\urldef\tempurl%
\url{https://doi.org/10.1177/1059712319844182}
\showDOI{\tempurl}


\bibitem[\protect\citeauthoryear{Andrist, Mutlu, and Tapus}{Andrist
  et~al\mbox{.}}{2015}]%
        {Andrist2015}
\bibfield{author}{\bibinfo{person}{Sean Andrist}, \bibinfo{person}{Bilge
  Mutlu}, {and} \bibinfo{person}{Adriana Tapus}.}
  \bibinfo{year}{2015}\natexlab{}.
\newblock \showarticletitle{{Look like me: Matching robot personality via gaze
  to increase motivation}}. In \bibinfo{booktitle}{\emph{Conference on Human
  Factors in Computing Systems - Proceedings}},
  Vol.~\bibinfo{volume}{2015-April}. \bibinfo{publisher}{Association for
  Computing Machinery}, \bibinfo{pages}{3603--3612}.
\newblock
\showISBNx{9781450331456}
\urldef\tempurl%
\url{https://doi.org/10.1145/2702123.2702592}
\showDOI{\tempurl}


\bibitem[\protect\citeauthoryear{Anjomshoae, Najjar, Calvaresi, and
  Fr{\"a}mling}{Anjomshoae et~al\mbox{.}}{2019}]%
        {anjomshoae2019explainable}
\bibfield{author}{\bibinfo{person}{Sule Anjomshoae}, \bibinfo{person}{Amro
  Najjar}, \bibinfo{person}{Davide Calvaresi}, {and} \bibinfo{person}{Kary
  Fr{\"a}mling}.} \bibinfo{year}{2019}\natexlab{}.
\newblock \showarticletitle{Explainable agents and robots: Results from a
  systematic literature review}. In \bibinfo{booktitle}{\emph{Proceedings of
  the 18th International Conference on Autonomous Agents and MultiAgent
  Systems}}. International Foundation for Autonomous Agents and Multiagent
  Systems, \bibinfo{pages}{1078--1088}.
\newblock


\bibitem[\protect\citeauthoryear{Bartneck, Kanda, Mubin, and {Al
  Mahmud}}{Bartneck et~al\mbox{.}}{2009a}]%
        {Bartneck2009}
\bibfield{author}{\bibinfo{person}{Christoph Bartneck},
  \bibinfo{person}{Takayuki Kanda}, \bibinfo{person}{Omar Mubin}, {and}
  \bibinfo{person}{Abdullah {Al Mahmud}}.} \bibinfo{year}{2009}\natexlab{a}.
\newblock \showarticletitle{{Does the Design of a Robot Influence Its Animacy
  and Perceived Intelligence?}}
\newblock \bibinfo{journal}{\emph{International Journal of Social Robotics}}
  \bibinfo{volume}{1}, \bibinfo{number}{2} (\bibinfo{date}{apr}
  \bibinfo{year}{2009}), \bibinfo{pages}{195--204}.
\newblock
\showISBNx{1875-4805 (Electronic)$\backslash$r1875-4791 (Print)}
\showISSN{1875-4791}
\urldef\tempurl%
\url{https://doi.org/10.1007/s12369-009-0013-7}
\showDOI{\tempurl}


\bibitem[\protect\citeauthoryear{Bartneck, Kuli{\'{c}}, Croft, and
  Zoghbi}{Bartneck et~al\mbox{.}}{2009b}]%
        {Bartneck2009a}
\bibfield{author}{\bibinfo{person}{Christoph Bartneck}, \bibinfo{person}{Dana
  Kuli{\'{c}}}, \bibinfo{person}{Elizabeth Croft}, {and}
  \bibinfo{person}{Susana Zoghbi}.} \bibinfo{year}{2009}\natexlab{b}.
\newblock \showarticletitle{{Measurement instruments for the anthropomorphism,
  animacy, likeability, perceived intelligence, and perceived safety of
  robots}}.
\newblock \bibinfo{journal}{\emph{International Journal of Social Robotics}}
  \bibinfo{volume}{1}, \bibinfo{number}{1} (\bibinfo{year}{2009}),
  \bibinfo{pages}{71--81}.
\newblock
\showISSN{18754791}
\urldef\tempurl%
\url{https://doi.org/10.1007/s12369-008-0001-3}
\showDOI{\tempurl}


\bibitem[\protect\citeauthoryear{Beer, Fisk, and Rogers}{Beer
  et~al\mbox{.}}{2014}]%
        {Beer2014}
\bibfield{author}{\bibinfo{person}{Jenay~M Beer}, \bibinfo{person}{Arthur~D
  Fisk}, {and} \bibinfo{person}{Wendy~A Rogers}.}
  \bibinfo{year}{2014}\natexlab{}.
\newblock \showarticletitle{{Toward a Framework for Levels of Robot Autonomy in
  Human-Robot Interaction}}.
\newblock \bibinfo{journal}{\emph{Journal of Human-Robot Interaction}}
  \bibinfo{volume}{3}, \bibinfo{number}{2} (\bibinfo{date}{jun}
  \bibinfo{year}{2014}), \bibinfo{pages}{74}.
\newblock
\showISSN{2163-0364}
\urldef\tempurl%
\url{https://doi.org/10.5898/jhri.3.2.beer}
\showDOI{\tempurl}


\bibitem[\protect\citeauthoryear{Breazeal, Kidd, Thomaz, Hoffman, and
  Berlin}{Breazeal et~al\mbox{.}}{2005}]%
        {Breazeal2005}
\bibfield{author}{\bibinfo{person}{Cynthia Breazeal}, \bibinfo{person}{Cory~D.
  Kidd}, \bibinfo{person}{Andrea~L. Thomaz}, \bibinfo{person}{Guy Hoffman},
  {and} \bibinfo{person}{Matt Berlin}.} \bibinfo{year}{2005}\natexlab{}.
\newblock \showarticletitle{{Effects of nonverbal communication on efficiency
  and robustness in human-robot teamwork}}. In \bibinfo{booktitle}{\emph{2005
  IEEE/RSJ International Conference on Intelligent Robots and Systems, IROS}}.
  \bibinfo{pages}{383--388}.
\newblock
\showISBNx{0780389123}
\urldef\tempurl%
\url{https://doi.org/10.1109/IROS.2005.1545011}
\showDOI{\tempurl}


\bibitem[\protect\citeauthoryear{Broadbent, Kumar, Li, Sollers, Stafford,
  MacDonald, and Wegner}{Broadbent et~al\mbox{.}}{2013}]%
        {Broadbent2013}
\bibfield{author}{\bibinfo{person}{Elizabeth Broadbent},
  \bibinfo{person}{Vinayak Kumar}, \bibinfo{person}{Xingyan Li},
  \bibinfo{person}{John Sollers}, \bibinfo{person}{Rebecca~Q. Stafford},
  \bibinfo{person}{Bruce~A. MacDonald}, {and} \bibinfo{person}{Daniel~M.
  Wegner}.} \bibinfo{year}{2013}\natexlab{}.
\newblock \showarticletitle{{Robots with Display Screens: A Robot with a More
  Humanlike Face Display Is Perceived To Have More Mind and a Better
  Personality}}.
\newblock \bibinfo{journal}{\emph{PLoS ONE}} \bibinfo{volume}{8},
  \bibinfo{number}{8} (\bibinfo{date}{aug} \bibinfo{year}{2013}),
  \bibinfo{pages}{e72589}.
\newblock
\showISBNx{1932-6203}
\showISSN{1932-6203}
\urldef\tempurl%
\url{https://doi.org/10.1371/journal.pone.0072589}
\showDOI{\tempurl}


\bibitem[\protect\citeauthoryear{Caruana, Spirou, and Brock}{Caruana
  et~al\mbox{.}}{2017}]%
        {Caruana2017}
\bibfield{author}{\bibinfo{person}{Nathan Caruana}, \bibinfo{person}{Dean
  Spirou}, {and} \bibinfo{person}{Jon Brock}.} \bibinfo{year}{2017}\natexlab{}.
\newblock \showarticletitle{Human agency beliefs influence behaviour during
  virtual social interactions}. In \bibinfo{booktitle}{\emph{PeerJ}}.
\newblock


\bibitem[\protect\citeauthoryear{Cohen}{Cohen}{1992}]%
        {Cohen1992}
\bibfield{author}{\bibinfo{person}{Jacob Cohen}.}
  \bibinfo{year}{1992}\natexlab{}.
\newblock \showarticletitle{{Quantitative methods in psychology}}.
\newblock \bibinfo{journal}{\emph{Nature}} \bibinfo{volume}{141},
  \bibinfo{number}{3570} (\bibinfo{year}{1992}), \bibinfo{pages}{613}.
\newblock
\showISSN{00280836}
\urldef\tempurl%
\url{https://doi.org/10.1038/141613a0}
\showDOI{\tempurl}


\bibitem[\protect\citeauthoryear{Corrigan, Peters, and Castellano}{Corrigan
  et~al\mbox{.}}{2013}]%
        {Corrigan2013}
\bibfield{author}{\bibinfo{person}{Lee~J. Corrigan},
  \bibinfo{person}{Christopher Peters}, {and} \bibinfo{person}{Ginevra
  Castellano}.} \bibinfo{year}{2013}\natexlab{}.
\newblock \showarticletitle{{Identifying task engagement: Towards personalised
  interactions with educational robots}}. In
  \bibinfo{booktitle}{\emph{Proceedings - 2013 Humaine Association Conference
  on Affective Computing and Intelligent Interaction, ACII 2013}}.
  \bibinfo{publisher}{IEEE}, \bibinfo{pages}{655--658}.
\newblock
\showISBNx{9780769550480}
\urldef\tempurl%
\url{https://doi.org/10.1109/ACII.2013.114}
\showDOI{\tempurl}


\bibitem[\protect\citeauthoryear{Deng, Mutlu, and Mataric}{Deng
  et~al\mbox{.}}{2019}]%
        {Deng2019}
\bibfield{author}{\bibinfo{person}{Eric Deng}, \bibinfo{person}{Bilge Mutlu},
  {and} \bibinfo{person}{Maja~J. Mataric}.} \bibinfo{year}{2019}\natexlab{}.
\newblock \showarticletitle{{Embodiment in Socially Interactive Robots}}.
\newblock \bibinfo{journal}{\emph{Foundations and Trends in Robotics}}
  \bibinfo{volume}{7}, \bibinfo{number}{4} (\bibinfo{year}{2019}),
  \bibinfo{pages}{251--356}.
\newblock
\showISBNx{9781680835465}
\showISSN{1935-8253}
\urldef\tempurl%
\url{https://doi.org/10.1561/2300000056}
\showDOI{\tempurl}


\bibitem[\protect\citeauthoryear{Epley, Waytz, and Cacioppo}{Epley
  et~al\mbox{.}}{2007}]%
        {Epley2007}
\bibfield{author}{\bibinfo{person}{Nicholas Epley}, \bibinfo{person}{Adam
  Waytz}, {and} \bibinfo{person}{John~T. Cacioppo}.}
  \bibinfo{year}{2007}\natexlab{}.
\newblock \showarticletitle{{On seeing human: A three-factor theory of
  anthropomorphism.}}
\newblock \bibinfo{journal}{\emph{Psychological Review}} \bibinfo{volume}{114},
  \bibinfo{number}{4} (\bibinfo{year}{2007}), \bibinfo{pages}{864--886}.
\newblock
\showISBNx{0033-295X (Print)$\backslash$r0033-295X (Linking)}
\showISSN{1939-1471}
\urldef\tempurl%
\url{https://doi.org/10.1037/0033-295X.114.4.864}
\showDOI{\tempurl}
\showeprint[arxiv]{epley2007}


\bibitem[\protect\citeauthoryear{Eyssel, Kuchenbrandt, Bobinger, de~Ruiter, and
  Hegel}{Eyssel et~al\mbox{.}}{2012}]%
        {Eyssel2012}
\bibfield{author}{\bibinfo{person}{Friederike~A. Eyssel},
  \bibinfo{person}{Dieta Kuchenbrandt}, \bibinfo{person}{Simon Bobinger},
  \bibinfo{person}{Laura de Ruiter}, {and} \bibinfo{person}{Frank Hegel}.}
  \bibinfo{year}{2012}\natexlab{}.
\newblock \showarticletitle{{'If you sound like me, you must be more human'}}.
  In \bibinfo{booktitle}{\emph{Proceedings of the seventh annual ACM/IEEE
  International Conference on Human-Robot Interaction - HRI '12}}.
  \bibinfo{publisher}{ACM Press}, \bibinfo{address}{New York, New York, USA},
  \bibinfo{pages}{125}.
\newblock
\showISBNx{9781450310635}
\showISSN{2167-2121}
\urldef\tempurl%
\url{https://doi.org/10.1145/2157689.2157717}
\showDOI{\tempurl}


\bibitem[\protect\citeauthoryear{Faul, Erdfelder, Lang, and Buchner}{Faul
  et~al\mbox{.}}{2007}]%
        {Faul2007}
\bibfield{author}{\bibinfo{person}{Franz Faul}, \bibinfo{person}{Edgar
  Erdfelder}, \bibinfo{person}{Albert~G. Lang}, {and} \bibinfo{person}{Axel
  Buchner}.} \bibinfo{year}{2007}\natexlab{}.
\newblock \showarticletitle{{G*Power 3: A flexible statistical power analysis
  program for the social, behavioral, and biomedical sciences}}.
\newblock \bibinfo{journal}{\emph{Behavior Research Methods}}
  \bibinfo{volume}{39}, \bibinfo{number}{2} (\bibinfo{year}{2007}),
  \bibinfo{pages}{175--191}.
\newblock
\showISSN{1554351X}
\urldef\tempurl%
\url{https://doi.org/10.3758/BF03193146}
\showDOI{\tempurl}


\bibitem[\protect\citeauthoryear{Fink}{Fink}{2012}]%
        {Fink2012}
\bibfield{author}{\bibinfo{person}{Julia Fink}.}
  \bibinfo{year}{2012}\natexlab{}.
\newblock \showarticletitle{Anthropomorphism and Human Likeness in the Design
  of Robots and Human-Robot Interaction}. In
  \bibinfo{booktitle}{\emph{International Conference on Social Robotics}}.
  \bibinfo{pages}{199--208}.
\newblock


\bibitem[\protect\citeauthoryear{Fischer, Lohan, Saunders, Nehaniv, Wrede, and
  Rohlfing}{Fischer et~al\mbox{.}}{2013}]%
        {Fischer2013}
\bibfield{author}{\bibinfo{person}{Kerstin Fischer}, \bibinfo{person}{Katrin
  Lohan}, \bibinfo{person}{Joe Saunders}, \bibinfo{person}{Chrystopher
  Nehaniv}, \bibinfo{person}{Britta Wrede}, {and} \bibinfo{person}{Katharina
  Rohlfing}.} \bibinfo{year}{2013}\natexlab{}.
\newblock \showarticletitle{{The impact of the contingency of robot feedback on
  HRI}}. In \bibinfo{booktitle}{\emph{Proceedings of the 2013 International
  Conference on Collaboration Technologies and Systems, CTS 2013}}.
  \bibinfo{publisher}{IEEE}, \bibinfo{pages}{210--217}.
\newblock
\showISBNx{9781467364027}
\urldef\tempurl%
\url{https://doi.org/10.1109/CTS.2013.6567231}
\showDOI{\tempurl}


\bibitem[\protect\citeauthoryear{Ghazali, Ham, Markopoulos, and
  Barakova}{Ghazali et~al\mbox{.}}{2019}]%
        {Ghazali2019a}
\bibfield{author}{\bibinfo{person}{Aimi~S. Ghazali}, \bibinfo{person}{Jaap
  Ham}, \bibinfo{person}{Panos Markopoulos}, {and} \bibinfo{person}{Emilia
  Barakova}.} \bibinfo{year}{2019}\natexlab{}.
\newblock \showarticletitle{{Investigating the Effect of Social Cues on Social
  Agency Judgement}}. In \bibinfo{booktitle}{\emph{2019 14th ACM/IEEE
  International Conference on Human-Robot Interaction (HRI)}}.
  \bibinfo{publisher}{IEEE}, \bibinfo{pages}{586--587}.
\newblock
\showISBNx{978-1-5386-8555-6}
\urldef\tempurl%
\url{https://doi.org/10.1109/HRI.2019.8673266}
\showDOI{\tempurl}


\bibitem[\protect\citeauthoryear{Graaf}{Graaf}{2019}]%
        {Graaf2019}
\bibfield{author}{\bibinfo{person}{Maartje M. A.~De Graaf}.}
  \bibinfo{year}{2019}\natexlab{}.
\newblock \showarticletitle{{People's Explanations of Robot Behavior Subtly
  Reveal Mental State Inferences}}.
\newblock \bibinfo{journal}{\emph{2019 14th ACM/IEEE International Conference
  on Human-Robot Interaction (HRI)}}, \bibinfo{pages}{239--248}.
\newblock
\showISBNx{9781538685556}


\bibitem[\protect\citeauthoryear{Gray, Gray, and Wegner}{Gray
  et~al\mbox{.}}{2007}]%
        {Gray2007}
\bibfield{author}{\bibinfo{person}{Heather~M. Gray}, \bibinfo{person}{Kurt
  Gray}, {and} \bibinfo{person}{Daniel~M. Wegner}.}
  \bibinfo{year}{2007}\natexlab{}.
\newblock \showarticletitle{{Dimensions of Mind Perception}}.
\newblock \bibinfo{journal}{\emph{Science}} \bibinfo{volume}{315},
  \bibinfo{number}{5812} (\bibinfo{date}{feb} \bibinfo{year}{2007}),
  \bibinfo{pages}{619--619}.
\newblock
\showISBNx{0036-8075}
\showISSN{0036-8075}
\urldef\tempurl%
\url{https://doi.org/10.1126/science.1134475}
\showDOI{\tempurl}


\bibitem[\protect\citeauthoryear{Groom, Srinivasan, Bethel, Murphy, Dole, and
  Nass}{Groom et~al\mbox{.}}{2011}]%
        {Groom2011}
\bibfield{author}{\bibinfo{person}{Victoria Groom}, \bibinfo{person}{Vasant
  Srinivasan}, \bibinfo{person}{Cindy~L. Bethel}, \bibinfo{person}{Robin
  Murphy}, \bibinfo{person}{Lorin Dole}, {and} \bibinfo{person}{Clifford
  Nass}.} \bibinfo{year}{2011}\natexlab{}.
\newblock \showarticletitle{{Responses to robot social roles and social role
  framing}}. In \bibinfo{booktitle}{\emph{2011 International Conference on
  Collaboration Technologies and Systems (CTS)}}. \bibinfo{publisher}{IEEE},
  \bibinfo{pages}{194--203}.
\newblock
\showISBNx{978-1-61284-638-5}
\urldef\tempurl%
\url{https://doi.org/10.1109/CTS.2011.5928687}
\showDOI{\tempurl}


\bibitem[\protect\citeauthoryear{Hoffmann and Kr{\"{a}}mer}{Hoffmann and
  Kr{\"{a}}mer}{2011}]%
        {Hoffmann2011}
\bibfield{author}{\bibinfo{person}{Laura Hoffmann} {and}
  \bibinfo{person}{Nicole~C. Kr{\"{a}}mer}.} \bibinfo{year}{2011}\natexlab{}.
\newblock \showarticletitle{{How should an artificial entity be embodied?
  Comparing the effects of a physically present robot and its virtual
  representation}}.
\newblock \bibinfo{journal}{\emph{HRI 2011 workshop on social robotic
  telepresence}} (\bibinfo{year}{2011}), \bibinfo{pages}{14--20}.
\newblock


\bibitem[\protect\citeauthoryear{Irfan, Ramachandran, Spaulding, Glas, Leite,
  and Koay}{Irfan et~al\mbox{.}}{2019}]%
        {Irfan2019}
\bibfield{author}{\bibinfo{person}{Bahar Irfan}, \bibinfo{person}{Aditi
  Ramachandran}, \bibinfo{person}{Samuel Spaulding}, \bibinfo{person}{Dylan~F.
  Glas}, \bibinfo{person}{Iolanda Leite}, {and} \bibinfo{person}{Kheng~L.
  Koay}.} \bibinfo{year}{2019}\natexlab{}.
\newblock \showarticletitle{{Personalization in Long-Term Human-Robot
  Interaction}}. In \bibinfo{booktitle}{\emph{ACM/IEEE International Conference
  on Human-Robot Interaction}}, Vol.~\bibinfo{volume}{2019-March}.
  \bibinfo{publisher}{IEEE}, \bibinfo{pages}{685--686}.
\newblock
\showISBNx{9781538685556}
\showISSN{21672148}
\urldef\tempurl%
\url{https://doi.org/10.1109/HRI.2019.8673076}
\showDOI{\tempurl}


\bibitem[\protect\citeauthoryear{Ishii and Watanabe}{Ishii and
  Watanabe}{2019}]%
        {Ishii2019}
\bibfield{author}{\bibinfo{person}{Tatsunori Ishii} {and}
  \bibinfo{person}{Katsumi Watanabe}.} \bibinfo{year}{2019}\natexlab{}.
\newblock \showarticletitle{{How People Attribute Minds to Non-Living
  Entities}}. In \bibinfo{booktitle}{\emph{2019 11th International Conference
  on Knowledge and Smart Technology, KST 2019}}. \bibinfo{publisher}{Institute
  of Electrical and Electronics Engineers Inc.}, \bibinfo{pages}{213--217}.
\newblock
\showISBNx{9781538675120}
\urldef\tempurl%
\url{https://doi.org/10.1109/KST.2019.8687324}
\showDOI{\tempurl}


\bibitem[\protect\citeauthoryear{Jung, Lee, Depalma, Adalgeirsson, Hinds, and
  Breazeal}{Jung et~al\mbox{.}}{2013}]%
        {Jung2013}
\bibfield{author}{\bibinfo{person}{Malte~F. Jung}, \bibinfo{person}{Jin~J.
  Lee}, \bibinfo{person}{Nick Depalma}, \bibinfo{person}{Sigurdur~O.
  Adalgeirsson}, \bibinfo{person}{Pamela~J. Hinds}, {and}
  \bibinfo{person}{Cynthia Breazeal}.} \bibinfo{year}{2013}\natexlab{}.
\newblock \showarticletitle{{Engaging robots: Easing complex human-robot
  teamwork using backchanneling}}.
\newblock \bibinfo{journal}{\emph{Proceedings of the ACM Conference on Computer
  Supported Cooperative Work, CSCW}}, \bibinfo{pages}{1555--1566}.
\newblock
\showISBNx{9781450313315}
\urldef\tempurl%
\url{https://doi.org/10.1145/2441776.2441954}
\showDOI{\tempurl}


\bibitem[\protect\citeauthoryear{Koda, Nishimura, and Nishijima}{Koda
  et~al\mbox{.}}{2016}]%
        {Koda2016}
\bibfield{author}{\bibinfo{person}{Tomoko Koda}, \bibinfo{person}{Yuta
  Nishimura}, {and} \bibinfo{person}{Tomofumi Nishijima}.}
  \bibinfo{year}{2016}\natexlab{}.
\newblock \showarticletitle{{How robot's animacy affects human tolerance for
  their malfunctions?}}. In \bibinfo{booktitle}{\emph{ACM/IEEE International
  Conference on Human-Robot Interaction}}, Vol.~\bibinfo{volume}{2016-April}.
  \bibinfo{publisher}{IEEE}, \bibinfo{pages}{455--456}.
\newblock
\showISBNx{9781467383707}
\showISSN{21672148}
\urldef\tempurl%
\url{https://doi.org/10.1109/HRI.2016.7451803}
\showDOI{\tempurl}


\bibitem[\protect\citeauthoryear{Kose-Bagci, Dautenhahn, and
  Nehaniv}{Kose-Bagci et~al\mbox{.}}{2008}]%
        {Kose-Bagci2008}
\bibfield{author}{\bibinfo{person}{Hatice Kose-Bagci}, \bibinfo{person}{Kerstin
  Dautenhahn}, {and} \bibinfo{person}{Chrystopher~L. Nehaniv}.}
  \bibinfo{year}{2008}\natexlab{}.
\newblock \showarticletitle{{Emergent dynamics of turn-taking interaction in
  drumming games with a humanoid robot}}. In \bibinfo{booktitle}{\emph{RO-MAN
  2008 - The 17th IEEE International Symposium on Robot and Human Interactive
  Communication}}. \bibinfo{publisher}{IEEE}, \bibinfo{pages}{346--353}.
\newblock
\showISBNx{978-1-4244-2212-8}
\urldef\tempurl%
\url{https://doi.org/10.1109/ROMAN.2008.4600690}
\showDOI{\tempurl}


\bibitem[\protect\citeauthoryear{Kwon, Jung, and Knepper}{Kwon
  et~al\mbox{.}}{2016}]%
        {Kwon2016}
\bibfield{author}{\bibinfo{person}{Minae Kwon}, \bibinfo{person}{Malte~F.
  Jung}, {and} \bibinfo{person}{Ross~A. Knepper}.}
  \bibinfo{year}{2016}\natexlab{}.
\newblock \showarticletitle{{Human expectations of social robots}}. In
  \bibinfo{booktitle}{\emph{2016 11th ACM/IEEE International Conference on
  Human-Robot Interaction (HRI)}}, Vol.~\bibinfo{volume}{2016-April}.
  \bibinfo{publisher}{IEEE}, \bibinfo{pages}{463--464}.
\newblock
\showISBNx{978-1-4673-8370-7}
\showISSN{21672148}
\urldef\tempurl%
\url{https://doi.org/10.1109/HRI.2016.7451807}
\showDOI{\tempurl}


\bibitem[\protect\citeauthoryear{Laham}{Laham}{2009}]%
        {Laham2009}
\bibfield{author}{\bibinfo{person}{Simon~M. Laham}.}
  \bibinfo{year}{2009}\natexlab{}.
\newblock \showarticletitle{{Expanding the moral circle: Inclusion and
  exclusion mindsets and the circle of moral regard}}.
\newblock \bibinfo{journal}{\emph{Journal of Experimental Social Psychology}}
  \bibinfo{volume}{45}, \bibinfo{number}{1} (\bibinfo{year}{2009}),
  \bibinfo{pages}{250--253}.
\newblock
\showISSN{00221031}
\urldef\tempurl%
\url{https://doi.org/10.1016/j.jesp.2008.08.012}
\showDOI{\tempurl}


\bibitem[\protect\citeauthoryear{Ligthart and Truong}{Ligthart and
  Truong}{2015}]%
        {Ligthart2015}
\bibfield{author}{\bibinfo{person}{Mike Ligthart} {and}
  \bibinfo{person}{Khiet~P. Truong}.} \bibinfo{year}{2015}\natexlab{}.
\newblock \showarticletitle{{Selecting the right robot: Influence of user
  attitude, robot sociability and embodiment on user preferences}}.
\newblock \bibinfo{journal}{\emph{Proceedings - IEEE International Workshop on
  Robot and Human Interactive Communication}}  \bibinfo{volume}{2015-Novem}
  (\bibinfo{year}{2015}), \bibinfo{pages}{682--687}.
\newblock
\showISBNx{9781467367042}
\urldef\tempurl%
\url{https://doi.org/10.1109/ROMAN.2015.7333598}
\showDOI{\tempurl}


\bibitem[\protect\citeauthoryear{Malle, Scheutz, Forlizzi, and Voiklis}{Malle
  et~al\mbox{.}}{2016}]%
        {Malle2016}
\bibfield{author}{\bibinfo{person}{Bertram~F. Malle}, \bibinfo{person}{Matthias
  Scheutz}, \bibinfo{person}{Jodi Forlizzi}, {and} \bibinfo{person}{John
  Voiklis}.} \bibinfo{year}{2016}\natexlab{}.
\newblock \showarticletitle{{Which robot am I thinking about? The impact of
  action and appearance on people's evaluations of a moral robot}}. In
  \bibinfo{booktitle}{\emph{ACM/IEEE International Conference on Human-Robot
  Interaction}}, Vol.~\bibinfo{volume}{2016-April}. \bibinfo{publisher}{IEEE
  Computer Society}, \bibinfo{pages}{125--132}.
\newblock
\showISBNx{9781467383707}
\showISSN{21672148}
\urldef\tempurl%
\url{https://doi.org/10.1109/HRI.2016.7451743}
\showDOI{\tempurl}


\bibitem[\protect\citeauthoryear{Marchesi, Ghiglino, Ciardo, Perez-Osorio,
  Baykara, and Wykowska}{Marchesi et~al\mbox{.}}{2019}]%
        {Marchesi2019}
\bibfield{author}{\bibinfo{person}{Serena Marchesi}, \bibinfo{person}{Davide
  Ghiglino}, \bibinfo{person}{Francesca Ciardo}, \bibinfo{person}{Jairo
  Perez-Osorio}, \bibinfo{person}{Ebru Baykara}, {and}
  \bibinfo{person}{Agnieszka Wykowska}.} \bibinfo{year}{2019}\natexlab{}.
\newblock \showarticletitle{Do We Adopt the Intentional Stance Toward Humanoid
  Robots?}
\newblock \bibinfo{journal}{\emph{Frontiers in Psychology}}
  \bibinfo{volume}{10} (\bibinfo{year}{2019}), \bibinfo{pages}{450}.
\newblock
\showISSN{1664-1078}
\urldef\tempurl%
\url{https://doi.org/10.3389/fpsyg.2019.00450}
\showDOI{\tempurl}


\bibitem[\protect\citeauthoryear{Mirnig, Stollnberger, Miksch, Stadler,
  Giuliani, and Tscheligi}{Mirnig et~al\mbox{.}}{2017}]%
        {Mirnig2017}
\bibfield{author}{\bibinfo{person}{Nicole Mirnig}, \bibinfo{person}{Gerald
  Stollnberger}, \bibinfo{person}{Markus Miksch}, \bibinfo{person}{Susanne
  Stadler}, \bibinfo{person}{Manuel Giuliani}, {and} \bibinfo{person}{Manfred
  Tscheligi}.} \bibinfo{year}{2017}\natexlab{}.
\newblock \showarticletitle{{To Err Is Robot: How Humans Assess and Act toward
  an Erroneous Social Robot}}.
\newblock \bibinfo{journal}{\emph{Frontiers in Robotics and AI}}
  \bibinfo{volume}{4} (\bibinfo{date}{may} \bibinfo{year}{2017}),
  \bibinfo{pages}{1--15}.
\newblock
\showISBNx{2296-9144}
\showISSN{2296-9144}
\urldef\tempurl%
\url{https://doi.org/10.3389/frobt.2017.00021}
\showDOI{\tempurl}


\bibitem[\protect\citeauthoryear{Mutlu, Forlizzi, and Hodgins}{Mutlu
  et~al\mbox{.}}{2006}]%
        {Mutlu2006}
\bibfield{author}{\bibinfo{person}{Bilge Mutlu}, \bibinfo{person}{Jodi
  Forlizzi}, {and} \bibinfo{person}{Jessica Hodgins}.}
  \bibinfo{year}{2006}\natexlab{}.
\newblock \showarticletitle{{A storytelling robot: Modeling and evaluation of
  human-like gaze behavior}}. In \bibinfo{booktitle}{\emph{Proceedings of the
  2006 6th IEEE-RAS International Conference on Humanoid Robots, HUMANOIDS}}.
  \bibinfo{pages}{518--523}.
\newblock
\showISBNx{142440200X}
\urldef\tempurl%
\url{https://doi.org/10.1109/ICHR.2006.321322}
\showDOI{\tempurl}


\bibitem[\protect\citeauthoryear{Nomura, Kanda, and Yamada}{Nomura
  et~al\mbox{.}}{2019a}]%
        {Nomura2019}
\bibfield{author}{\bibinfo{person}{Tatsuya Nomura}, \bibinfo{person}{Takayuki
  Kanda}, {and} \bibinfo{person}{Sachie Yamada}.}
  \bibinfo{year}{2019}\natexlab{a}.
\newblock \showarticletitle{{Measurement of Moral Concern for Robots}}. In
  \bibinfo{booktitle}{\emph{2019 14th ACM/IEEE International Conference on
  Human-Robot Interaction (HRI)}}. \bibinfo{publisher}{IEEE},
  \bibinfo{pages}{540--541}.
\newblock
\showISBNx{978-1-5386-8555-6}
\urldef\tempurl%
\url{https://doi.org/10.1109/HRI.2019.8673095}
\showDOI{\tempurl}


\bibitem[\protect\citeauthoryear{Nomura, Otsubo, and Kanda}{Nomura
  et~al\mbox{.}}{2019b}]%
        {Nomura2019a}
\bibfield{author}{\bibinfo{person}{Tatsuya Nomura}, \bibinfo{person}{Kazuki
  Otsubo}, {and} \bibinfo{person}{Takayuki Kanda}.}
  \bibinfo{year}{2019}\natexlab{b}.
\newblock \showarticletitle{{Preliminary Investigation of Moral Expansiveness
  for Robots}}.
\newblock \bibinfo{journal}{\emph{Proceedings of IEEE Workshop on Advanced
  Robotics and its Social Impacts, ARSO}}  \bibinfo{volume}{2018-September}
  (\bibinfo{year}{2019}), \bibinfo{pages}{91--96}.
\newblock
\showISBNx{9781538680377}
\showISSN{21627576}
\urldef\tempurl%
\url{https://doi.org/10.1109/ARSO.2018.8625717}
\showDOI{\tempurl}


\bibitem[\protect\citeauthoryear{Park, Kong, Lim, Lee, You, and {Del
  Pobil}}{Park et~al\mbox{.}}{2011}]%
        {Park2011}
\bibfield{author}{\bibinfo{person}{Eunil Park}, \bibinfo{person}{Hwayeon Kong},
  \bibinfo{person}{Hyeong~T. Lim}, \bibinfo{person}{Jongsik Lee},
  \bibinfo{person}{Sangseok You}, {and} \bibinfo{person}{Angel~P. {Del
  Pobil}}.} \bibinfo{year}{2011}\natexlab{}.
\newblock \showarticletitle{{The effect of robot's behavior vs. appearance on
  communication with humans}}. In \bibinfo{booktitle}{\emph{HRI 2011 -
  Proceedings of the 6th ACM/IEEE International Conference on Human-Robot
  Interaction}}. \bibinfo{publisher}{IEEE}, \bibinfo{pages}{219--220}.
\newblock
\showISBNx{9781450305617}
\urldef\tempurl%
\url{https://doi.org/10.1145/1957656.1957740}
\showDOI{\tempurl}


\bibitem[\protect\citeauthoryear{Phillips, Zhao, Ullman, and Malle}{Phillips
  et~al\mbox{.}}{2018}]%
        {Phillips2018}
\bibfield{author}{\bibinfo{person}{Elizabeth Phillips}, \bibinfo{person}{Xuan
  Zhao}, \bibinfo{person}{Daniel Ullman}, {and} \bibinfo{person}{Bertram~F.
  Malle}.} \bibinfo{year}{2018}\natexlab{}.
\newblock \showarticletitle{{What is Human-like?: Decomposing Robots'
  Human-like Appearance Using the Anthropomorphic roBOT (ABOT) Database}}. In
  \bibinfo{booktitle}{\emph{ACM/IEEE International Conference on Human-Robot
  Interaction}}. \bibinfo{publisher}{ACM Press}, \bibinfo{address}{New York,
  New York, USA}, \bibinfo{pages}{105--113}.
\newblock
\showISBNx{9781450349536}
\showISSN{21672148}
\urldef\tempurl%
\url{https://doi.org/10.1145/3171221.3171268}
\showDOI{\tempurl}


\bibitem[\protect\citeauthoryear{Rea and Young}{Rea and Young}{2018}]%
        {Rea2018}
\bibfield{author}{\bibinfo{person}{Daniel~J. Rea} {and}
  \bibinfo{person}{James~E. Young}.} \bibinfo{year}{2018}\natexlab{}.
\newblock \showarticletitle{{It's All in Your Head: Using Priming to Shape an
  Operator's Perceptions and Behavior during Teleoperation}}. In
  \bibinfo{booktitle}{\emph{Proceedings of the 2018 ACM/IEEE International
  Conference on Human-Robot Interaction - HRI '18}}. \bibinfo{pages}{32--40}.
\newblock
\showISBNx{9781450349536}
\urldef\tempurl%
\url{https://doi.org/10.1145/3171221.3171259}
\showDOI{\tempurl}


\bibitem[\protect\citeauthoryear{Rea and Young}{Rea and Young}{2019}]%
        {Rea2019}
\bibfield{author}{\bibinfo{person}{Daniel~J. Rea} {and}
  \bibinfo{person}{James~E. Young}.} \bibinfo{year}{2019}\natexlab{}.
\newblock \showarticletitle{{Methods and Effects of Priming a Teloperator's
  Perception of Robot Capabilities}}. In \bibinfo{booktitle}{\emph{ACM/IEEE
  International Conference on Human-Robot Interaction}},
  Vol.~\bibinfo{volume}{2019-March}. \bibinfo{publisher}{IEEE},
  \bibinfo{pages}{739--741}.
\newblock
\showISBNx{9781538685556}
\showISSN{21672148}
\urldef\tempurl%
\url{https://doi.org/10.1109/HRI.2019.8673186}
\showDOI{\tempurl}


\bibitem[\protect\citeauthoryear{Salem, Eyssel, Rohlfing, Kopp, and
  Joublin}{Salem et~al\mbox{.}}{2013}]%
        {Salem2013}
\bibfield{author}{\bibinfo{person}{Maha Salem}, \bibinfo{person}{Friederike~A.
  Eyssel}, \bibinfo{person}{Katharina Rohlfing}, \bibinfo{person}{Stefan Kopp},
  {and} \bibinfo{person}{Frank Joublin}.} \bibinfo{year}{2013}\natexlab{}.
\newblock \showarticletitle{{To Err is Human(-like): Effects of Robot Gesture
  on Perceived Anthropomorphism and Likability}}.
\newblock \bibinfo{journal}{\emph{International Journal of Social Robotics}}
  \bibinfo{volume}{5}, \bibinfo{number}{3} (\bibinfo{date}{aug}
  \bibinfo{year}{2013}), \bibinfo{pages}{313--323}.
\newblock
\showISBNx{1875-4791}
\showISSN{1875-4791}
\urldef\tempurl%
\url{https://doi.org/10.1007/s12369-013-0196-9}
\showDOI{\tempurl}


\bibitem[\protect\citeauthoryear{Schellen and Wykowska}{Schellen and
  Wykowska}{2019}]%
        {Schellen2019}
\bibfield{author}{\bibinfo{person}{Elef Schellen} {and}
  \bibinfo{person}{Agnieszka Wykowska}.} \bibinfo{year}{2019}\natexlab{}.
\newblock \showarticletitle{{Intentional mindset toward robots-open questions
  and methodological challenges}}.
\newblock \bibinfo{journal}{\emph{Frontiers in Robotics AI}}
  \bibinfo{volume}{6}, \bibinfo{number}{JAN} (\bibinfo{date}{jan}
  \bibinfo{year}{2019}).
\newblock
\showISSN{22969144}
\urldef\tempurl%
\url{https://doi.org/10.3389/frobt.2018.00139}
\showDOI{\tempurl}


\bibitem[\protect\citeauthoryear{Schneider and Kummert}{Schneider and
  Kummert}{2017}]%
        {Schneider2017}
\bibfield{author}{\bibinfo{person}{Sebastian Schneider} {and}
  \bibinfo{person}{Franz Kummert}.} \bibinfo{year}{2017}\natexlab{}.
\newblock \showarticletitle{{Does the user's evaluation of a socially assistive
  robot change based on presence and companionship type?}}. In
  \bibinfo{booktitle}{\emph{ACM/IEEE International Conference on Human-Robot
  Interaction}}. \bibinfo{pages}{277--278}.
\newblock
\showISBNx{9781450348850}
\showISSN{21672148}
\urldef\tempurl%
\url{https://doi.org/10.1145/3029798.3038418}
\showDOI{\tempurl}


\bibitem[\protect\citeauthoryear{Seo, Geiskkovitch, Nakane, King, and
  Young}{Seo et~al\mbox{.}}{2015}]%
        {Seo2015}
\bibfield{author}{\bibinfo{person}{Stela~H. Seo}, \bibinfo{person}{Denise
  Geiskkovitch}, \bibinfo{person}{Masayuki Nakane}, \bibinfo{person}{Corey
  King}, {and} \bibinfo{person}{James~E. Young}.}
  \bibinfo{year}{2015}\natexlab{}.
\newblock \showarticletitle{{Poor Thing! Would You Feel Sorry for a Simulated
  Robot?: A comparison of empathy toward a physical and a simulated robot}}.
\newblock \bibinfo{journal}{\emph{ACM/IEEE International Conference on
  Human-Robot Interaction}}  \bibinfo{volume}{2015-March},
  \bibinfo{pages}{125--132}.
\newblock
\showISBNx{9781450328821}
\showISSN{21672148}
\urldef\tempurl%
\url{https://doi.org/10.1145/2696454.2696471}
\showDOI{\tempurl}


\bibitem[\protect\citeauthoryear{Short, Hart, Vu, and Scassellati}{Short
  et~al\mbox{.}}{2010}]%
        {Short2010}
\bibfield{author}{\bibinfo{person}{Elaine Short}, \bibinfo{person}{Justin
  Hart}, \bibinfo{person}{Michelle Vu}, {and} \bibinfo{person}{Brian
  Scassellati}.} \bibinfo{year}{2010}\natexlab{}.
\newblock \showarticletitle{{No fair!! An interaction with a cheating robot}}.
  In \bibinfo{booktitle}{\emph{2010 5th ACM/IEEE International Conference on
  Human-Robot Interaction (HRI)}}. \bibinfo{publisher}{IEEE},
  \bibinfo{pages}{219--226}.
\newblock
\showISBNx{978-1-4244-4892-0}
\showISSN{978-1-4244-4892-0}
\urldef\tempurl%
\url{https://doi.org/10.1109/HRI.2010.5453193}
\showDOI{\tempurl}


\bibitem[\protect\citeauthoryear{Stenzel, Chinellato, Bou, del Pobil, Lappe,
  and Liepelt}{Stenzel et~al\mbox{.}}{2012}]%
        {Stenzel2012}
\bibfield{author}{\bibinfo{person}{Anna Stenzel}, \bibinfo{person}{Eris
  Chinellato}, \bibinfo{person}{Maria A.~T. Bou},
  \bibinfo{person}{{\'{A}}ngel~P. del Pobil}, \bibinfo{person}{Markus Lappe},
  {and} \bibinfo{person}{Roman Liepelt}.} \bibinfo{year}{2012}\natexlab{}.
\newblock \showarticletitle{{When humanoid robots become human-like interaction
  partners: Corepresentation of robotic actions.}}
\newblock \bibinfo{journal}{\emph{Journal of Experimental Psychology: Human
  Perception and Performance}} \bibinfo{volume}{38}, \bibinfo{number}{5}
  (\bibinfo{date}{oct} \bibinfo{year}{2012}), \bibinfo{pages}{1073--1077}.
\newblock
\showISSN{1939-1277}
\urldef\tempurl%
\url{https://doi.org/10.1037/a0029493}
\showDOI{\tempurl}


\bibitem[\protect\citeauthoryear{Thellman and Ziemke}{Thellman and
  Ziemke}{2017}]%
        {Thellman2017}
\bibfield{author}{\bibinfo{person}{Sam Thellman} {and} \bibinfo{person}{Tom
  Ziemke}.} \bibinfo{year}{2017}\natexlab{}.
\newblock \showarticletitle{{Social Attitudes Toward Robots are Easily
  Manipulated}}. In \bibinfo{booktitle}{\emph{Proceedings of the Companion of
  the 2017 ACM/IEEE International Conference on Human-Robot Interaction - HRI
  '17}}. \bibinfo{publisher}{ACM Press}, \bibinfo{address}{New York, New York,
  USA}, \bibinfo{pages}{299--300}.
\newblock
\showISBNx{9781450348850}
\urldef\tempurl%
\url{https://doi.org/10.1145/3029798.3038336}
\showDOI{\tempurl}


\bibitem[\protect\citeauthoryear{Tsiakas, Abujelala, Lioulemes, and
  Makedon}{Tsiakas et~al\mbox{.}}{2017}]%
        {Tsiakas2017}
\bibfield{author}{\bibinfo{person}{Konstantinos Tsiakas},
  \bibinfo{person}{Maher Abujelala}, \bibinfo{person}{Alexandros Lioulemes},
  {and} \bibinfo{person}{Fillia Makedon}.} \bibinfo{year}{2017}\natexlab{}.
\newblock \showarticletitle{{An intelligent Interactive Learning and Adaptation
  framework for robot-based vocational training}}.
\newblock \bibinfo{journal}{\emph{2016 IEEE Symposium Series on Computational
  Intelligence, SSCI 2016}} \bibinfo{number}{October} (\bibinfo{year}{2017}).
\newblock
\showISBNx{9781509042401}
\urldef\tempurl%
\url{https://doi.org/10.1109/SSCI.2016.7850066}
\showDOI{\tempurl}


\bibitem[\protect\citeauthoryear{Vanman and Kappas}{Vanman and Kappas}{2019}]%
        {Vanman2019}
\bibfield{author}{\bibinfo{person}{Eric~J. Vanman} {and} \bibinfo{person}{Arvid
  Kappas}.} \bibinfo{year}{2019}\natexlab{}.
\newblock \showarticletitle{{"Danger, Will Robinson!" The challenges of social
  robots for intergroup relations}}.
\newblock \bibinfo{journal}{\emph{Social and Personality Psychology Compass}}
  \bibinfo{volume}{13}, \bibinfo{number}{8} (\bibinfo{date}{aug}
  \bibinfo{year}{2019}), \bibinfo{pages}{1--13}.
\newblock
\showISSN{1751-9004}
\urldef\tempurl%
\url{https://doi.org/10.1111/spc3.12489}
\showDOI{\tempurl}


\bibitem[\protect\citeauthoryear{Westlund, Martinez, Archie, Das, and
  Breazeal}{Westlund et~al\mbox{.}}{2016}]%
        {Westlund2016}
\bibfield{author}{\bibinfo{person}{Jacqueline M.~K. Westlund},
  \bibinfo{person}{Marayna Martinez}, \bibinfo{person}{Maryam Archie},
  \bibinfo{person}{Madhurima Das}, {and} \bibinfo{person}{Cynthia Breazeal}.}
  \bibinfo{year}{2016}\natexlab{}.
\newblock \showarticletitle{{A study to measure the effect of framing a robot
  as a social agent or as a machine on children's social behavior}}. In
  \bibinfo{booktitle}{\emph{ACM/IEEE International Conference on Human-Robot
  Interaction}}, Vol.~\bibinfo{volume}{2016-April}. \bibinfo{publisher}{IEEE
  Computer Society}, \bibinfo{pages}{459--460}.
\newblock
\showISBNx{9781467383707}
\showISSN{21672148}
\urldef\tempurl%
\url{https://doi.org/10.1109/HRI.2016.7451805}
\showDOI{\tempurl}


\bibitem[\protect\citeauthoryear{Wiese, Metta, and Wykowska}{Wiese
  et~al\mbox{.}}{2017}]%
        {Wiese2017}
\bibfield{author}{\bibinfo{person}{Eva Wiese}, \bibinfo{person}{Giorgio Metta},
  {and} \bibinfo{person}{Agnieszka Wykowska}.} \bibinfo{year}{2017}\natexlab{}.
\newblock \showarticletitle{{Robots as intentional agents: Using
  neuroscientific methods to make robots appear more social}}.
\newblock \bibinfo{journal}{\emph{Frontiers in psychology}}
  \bibinfo{volume}{8}, \bibinfo{number}{OCT} (\bibinfo{date}{oct}
  \bibinfo{year}{2017}).
\newblock
\showISSN{16641078}
\urldef\tempurl%
\url{https://doi.org/10.3389/fpsyg.2017.01663}
\showDOI{\tempurl}


\bibitem[\protect\citeauthoryear{Wiese, Wykowska, Zwickel, and Mueller}{Wiese
  et~al\mbox{.}}{2012}]%
        {Wiese2012}
\bibfield{author}{\bibinfo{person}{Eva Wiese}, \bibinfo{person}{Agnieszka
  Wykowska}, \bibinfo{person}{Jan Zwickel}, {and} \bibinfo{person}{Hermann~J.
  Mueller}.} \bibinfo{year}{2012}\natexlab{}.
\newblock \showarticletitle{I See What You Mean: How Attentional Selection Is
  Shaped by Ascribing Intentions to Others}. In \bibinfo{booktitle}{\emph{PloS
  one}}.
\newblock


\bibitem[\protect\citeauthoryear{Wigdor, de~Greeff, Looije, and
  Neerincx}{Wigdor et~al\mbox{.}}{2016}]%
        {Wigdor2016}
\bibfield{author}{\bibinfo{person}{Noel Wigdor}, \bibinfo{person}{Joachim de
  Greeff}, \bibinfo{person}{Rosemarijn Looije}, {and} \bibinfo{person}{Mark~A.
  Neerincx}.} \bibinfo{year}{2016}\natexlab{}.
\newblock \showarticletitle{{How to improve human-robot interaction with
  Conversational Fillers}}. In \bibinfo{booktitle}{\emph{2016 25th IEEE
  International Symposium on Robot and Human Interactive Communication
  (RO-MAN)}}. \bibinfo{publisher}{IEEE}, \bibinfo{pages}{219--224}.
\newblock
\showISBNx{978-1-5090-3929-6}
\urldef\tempurl%
\url{https://doi.org/10.1109/ROMAN.2016.7745134}
\showDOI{\tempurl}


\bibitem[\protect\citeauthoryear{Wills, Baxter, Kennedy, Senft, and
  Belpaeme}{Wills et~al\mbox{.}}{2016}]%
        {Wills2016}
\bibfield{author}{\bibinfo{person}{Paul Wills}, \bibinfo{person}{Paul Baxter},
  \bibinfo{person}{James Kennedy}, \bibinfo{person}{Emmanuel Senft}, {and}
  \bibinfo{person}{Tony Belpaeme}.} \bibinfo{year}{2016}\natexlab{}.
\newblock \showarticletitle{{Socially contingent humanoid robot head behaviour
  results in increased charity donations}}. In
  \bibinfo{booktitle}{\emph{ACM/IEEE International Conference on Human-Robot
  Interaction}}, Vol.~\bibinfo{volume}{2016-April}. \bibinfo{publisher}{IEEE},
  \bibinfo{pages}{533--534}.
\newblock
\showISBNx{9781467383707}
\showISSN{21672148}
\urldef\tempurl%
\url{https://doi.org/10.1109/HRI.2016.7451842}
\showDOI{\tempurl}


\bibitem[\protect\citeauthoryear{Wykowska, Wiese, Prosser, and
  Mueller}{Wykowska et~al\mbox{.}}{2014}]%
        {Wykowska2014}
\bibfield{author}{\bibinfo{person}{Agnieszka Wykowska}, \bibinfo{person}{Eva
  Wiese}, \bibinfo{person}{Aaron Prosser}, {and} \bibinfo{person}{Hermann~J.
  Mueller}.} \bibinfo{year}{2014}\natexlab{}.
\newblock \showarticletitle{Beliefs about the Minds of Others Influence How We
  Process Sensory Information}. In \bibinfo{booktitle}{\emph{PloS one}}.
\newblock


\bibitem[\protect\citeauthoryear{Z{\l}otowski, Proudfoot, Yogeeswaran, and
  Bartneck}{Z{\l}otowski et~al\mbox{.}}{2015}]%
        {Zlotowski2015}
\bibfield{author}{\bibinfo{person}{Jakub Z{\l}otowski}, \bibinfo{person}{Diane
  Proudfoot}, \bibinfo{person}{Kumar Yogeeswaran}, {and}
  \bibinfo{person}{Christoph Bartneck}.} \bibinfo{year}{2015}\natexlab{}.
\newblock \showarticletitle{{Anthropomorphism: Opportunities and Challenges in
  Human–Robot Interaction}}.
\newblock \bibinfo{journal}{\emph{International Journal of Social Robotics}}
  \bibinfo{volume}{7}, \bibinfo{number}{3} (\bibinfo{date}{jun}
  \bibinfo{year}{2015}), \bibinfo{pages}{347--360}.
\newblock
\showISBNx{1875-4791}
\showISSN{1875-4791}
\urldef\tempurl%
\url{https://doi.org/10.1007/s12369-014-0267-6}
\showDOI{\tempurl}


\end{thebibliography}


\end{document}